\documentclass[]{fairmeta}

\usepackage{amsmath,amsfonts,bm}

\usepackage{xspace}

\def\eqref#1{equation~\ref{#1}}

\def\1{\bm{1}}

\DeclareMathAlphabet{\mathsfit}{\encodingdefault}{\sfdefault}{m}{sl}
\SetMathAlphabet{\mathsfit}{bold}{\encodingdefault}{\sfdefault}{bx}{n}

\usepackage{url}            
\usepackage{amsfonts}       
\usepackage{nicefrac}       
\usepackage{xcolor}         
\usepackage{float}          

\usepackage{fvextra}
\usepackage{needspace}

\DefineVerbatimEnvironment{Prompt}{Verbatim}{
  breaklines=true,
  breakanywhere=true,
  fontsize=\scriptsize,
  frame=single,
  framesep=2mm
}

\usepackage{CJKutf8}

\usepackage{xcolor}
\usepackage{graphicx}
\usepackage{mdframed}
\usepackage{fvextra}
\usepackage{needspace}

\graphicspath{{omnimat_cases_overleaf/figures/omnimat_cases/}}

\definecolor{CaseUserHeader}{HTML}{526499}
\definecolor{CaseUserBody}{HTML}{FFFFFF}

\definecolor{CaseModelHeader}{HTML}{8A8DBC}
\definecolor{CaseModelBody}{HTML}{FFFFFF}

\definecolor{CaseBorder}{HTML}{D5DDE3}

\newmdenv[
  linewidth=0.6pt,
  linecolor=CaseUserHeader,
  backgroundcolor=CaseUserBody,
  roundcorner=4pt,
  frametitle={\strut User},
  frametitlebackgroundcolor=CaseUserHeader,
  frametitlefont=\color{white}\bfseries,
  frametitleaboveskip=4pt,
  frametitlebelowskip=4pt,
  innerleftmargin=10pt,
  innerrightmargin=10pt,
  innertopmargin=8pt,
  innerbottommargin=8pt,
  skipabove=7pt,
  skipbelow=7pt,
  nobreak=true
]{OmniUserBox}

\newmdenv[
  linewidth=0.6pt,
  linecolor=CaseModelHeader,
  backgroundcolor=CaseModelBody,
  roundcorner=4pt,
  frametitle={\strut MatVerse},
  frametitlebackgroundcolor=CaseModelHeader,
  frametitlefont=\color{white}\bfseries,
  frametitleaboveskip=4pt,
  frametitlebelowskip=4pt,
  innerleftmargin=10pt,
  innerrightmargin=10pt,
  innertopmargin=8pt,
  innerbottommargin=8pt,
  skipabove=7pt,
  skipbelow=12pt,
  nobreak=true
]{OmniModelBox}

\newcommand{\CaseTitle}[1]{%
  \par\medskip
  \noindent{\bfseries #1}\par
  \vspace{2pt}
}

\title{\resizebox{\linewidth}{!}{MatVerse: A Foundation Model for Materials Science}}

\newlength{\matverseAuthorWidth}
\AtBeginDocument{\settowidth{\matverseAuthorWidth}{\huge\sffamily\bfseries MatVerse: A Foundation Model for Materials}}

\author{
\vspace{0.6em}
\parbox{\matverseAuthorWidth}{%
\leftskip=0pt \rightskip=0pt \parfillskip=0pt plus 1fil
\mbox{Xuanyu Zhu$^{1,2,3,*}$},
\mbox{Qian Tan$^{2,4,*}$},
\mbox{Bo Chen$^{1}$},
\mbox{Jiaqing Xie$^{2,5}$},
\mbox{Yang Han$^{1,2,3}$},
\mbox{Mei Wu$^{1,3}$},
\mbox{Wanhao Liu$^{2,4}$},
\mbox{Xuanze Lin$^{3}$},
\mbox{Yuchen Zhou$^{3}$},
\mbox{Shijie Wang$^{2}$},
\mbox{Biqing Qi$^{2}$},
\mbox{Shixiang Tang$^{2}$},
\mbox{Wenwei Zhang$^{2}$},
\mbox{Tianfan Fu$^{6,2}$},
\mbox{Kai Yu$^{1,3}$},
\mbox{Lu Chen$^{1,3,\dagger}$},
\mbox{Yuqiang Li$^{2,\dagger}$},
\mbox{Xin Chen$^{1,\dagger}$}\par
}\\[0.8em]
{\small
$^{1}$ Suzhou Laboratory \\
$^{2}$ Shanghai Artificial Intelligence Laboratory \\
$^{3}$ Shanghai Jiao Tong University \\
$^{4}$ University of Science and Technology of China \\
$^{5}$ Fudan University \\
$^{6}$ Nanjing University
}\\[0.5em]
{\small
$^*$ These authors contributed equally. \\
$^\dagger$ Corresponding authors.
}
}

\abstract{Materials science involves highly heterogeneous scientific information,
including textual knowledge, molecular and crystal structures,
experimental characterization data, microscopy images, and
structure--property relationships. Developing a general-purpose foundation
model capable of understanding and reasoning over these diverse materials
representations remains challenging. In this technical report, we introduce MatVerse, a materials foundation
model built upon Intern-S2-Preview and further specialized through
large-scale materials-domain post-training. We construct a large-scale
multimodal training corpus covering materials knowledge, scientific
reasoning, quantitative problem solving, structural understanding, and
multimodal scientific interpretation. To improve domain adaptation, we
develop a unified data construction pipeline combining teacher-guided
reasoning distillation, scientific problem synthesis, consistency-based
filtering, and retrieval-based multimodal data construction. We evaluate MatVerse on three
representative materials-science benchmarks: our human-calibrated
OmniMatBench, MatSciBench, and MaCBench. Results demonstrate consistent
improvements after materials-specific post-training over the base model
on all three benchmarks. These results show that targeted
domain-specific post-training can effectively enhance scientific
reasoning, materials knowledge understanding, and multimodal scientific
capabilities of foundation models. MatVerse provides a unified language-and-vision interface for materials
science and serves as a foundation for future AI-assisted materials
research workflows.
}

\begin{document}
\begin{CJK*}{UTF8}{gbsn}

\maketitle

\newcommand{\method}{\textbf{MatVerse}}
\newcommand{\ESA}{exact slot accuracy}
\newcommand{\MacroFOne}{Macro-F$_1$}
\newcommand{\RuleAcc}{rule accuracy}
\newcommand{\OverAcc}{overall accuracy}

\begin{figure}[!ht]
    \centering
    \includegraphics[width=\textwidth]{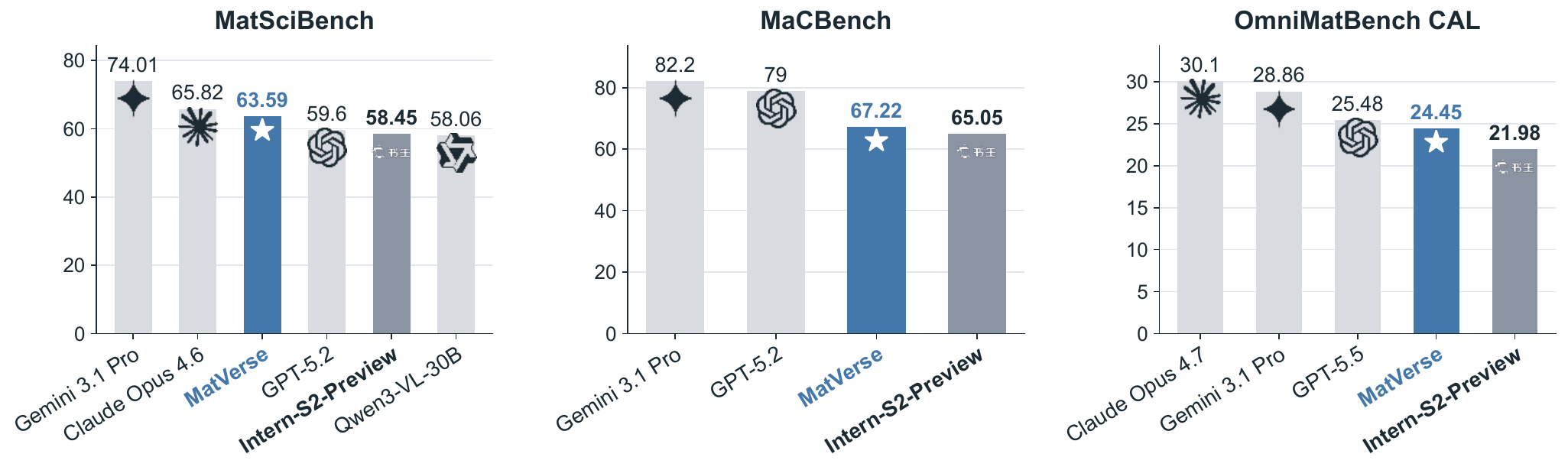}
    \caption{Overall comparison of \method{} against representative foundation models on MatSciBench, MaCBench, and the OmniMatBench CAL track. On every benchmark, \method{} substantially improves over its base model Intern-S2-Preview (35B) and achieves performance competitive with much larger frontier proprietary models, demonstrating that materials-specific post-training yields strong gains in scientific reasoning, multimodal scientific understanding, and quantitative calculation.}
    \label{fig:overall_ranking}
\end{figure}

\section{Introduction}
Materials science spans highly heterogeneous scientific information, ranging from symbolic and textual representations of chemical systems to molecular and crystal structures, experimental characterization signals, microstructural observations, and property-condition relationships~\citep{wan2024tokensmaterialsleveraginglanguage,alampara2025probinglimitationsmultimodallanguage,liu2026omnimatbenchhumancalibratedmultimodalreasoning}. Although recent large language models (LLMs) have achieved impressive general-purpose capabilities~\citep{qwenteam2026qwen35omnitechnicalreport,kimiteam2026kimik3openfrontier,deepseekai2026deepseekv4highlyefficientmilliontoken,anthropic2026claudesonnet55,openai2026gpt6astra}, they often struggle to jointly interpret symbolic descriptions, structural representations, experimental observations, and quantitative relationships within a single workflow. Strengthening scientific foundation models with materials-domain knowledge, quantitative reasoning, and multimodal scientific understanding is therefore essential for building a unified backbone that supports the full materials-research workflow.

Meanwhile, existing materials-oriented multimodal LLMs are often developed for specific tasks, modalities, or narrow application scenarios~\citep{wan2024tokensmaterialsleveraginglanguage,zhao2024chemdfm,zhao2025developing,zhang2024chemllm,li2025chemvlm}. For example, chemistry-oriented LLMs such as ChemDFM, ChemDFM-X, and ChemLLM concentrate on molecular strings, reactions, and chemical text~\citep{zhao2024chemdfm,zhao2025developing,zhang2024chemllm}, while chemistry-focused VLMs such as ChemVLM emphasize molecular recognition and chemical reasoning over molecular imagery~\citep{li2025chemvlm}. Task-oriented pipelines such as LLM4Mat further tailor language models to individual downstream applications rather than serving as reusable backbones~\citep{wan2024tokensmaterialsleveraginglanguage}. As a result, several capability gaps remain. First, support for heterogeneous scientific visual content, including scientific tables, plots, diffraction patterns, and microscopy images, is limited beyond molecular representations. Second, quantitative calculation and structured scientific reasoning, which are central to materials research, are often addressed only within narrow task formats. Third, because these models are trained and evaluated on disconnected task distributions, their capabilities rarely transfer across knowledge-intensive question answering, quantitative problem solving, and multimodal scientific understanding. Moreover, inference-time cost and deployment efficiency are becoming first-order considerations for practical materials-research workflows. A compact domain-specialized model with a modest number of active parameters offers an attractive alternative to substantially larger general-purpose systems, provided that it preserves competitive domain performance. A unified foundation model capable of supporting this full spectrum of capabilities in an efficient and deployable form therefore remains underexplored. This motivates us to develop \method{}, a general-purpose materials foundation model designed to provide a common backbone for diverse materials-science tasks.

We introduce \method{}, a materials foundation model built upon Intern-S2-Preview~\citep{bai2026intern} and further specialized through large-scale domain-specific post-training. To enhance its capabilities for materials and molecular science, we construct approximately 150K materials-domain post-training examples that combine textual reasoning data with multimodal scientific content, covering fundamental materials knowledge, structure--property relationships, processing and manufacturing, and failure mechanisms, with representative scenarios involving diffraction and spectroscopy data, microscopy images, molecular and crystal structures, experimental equipment, scientific tables, and patent figures. The training recipe is validated on a smaller pilot model before being scaled to the final 35B backbone. Through this domain-oriented post-training, \method{} strengthens the base model in materials knowledge, quantitative reasoning, and multimodal scientific understanding.

We evaluate \method{} on three complementary materials-science benchmarks:
our human-calibrated
\textbf{OmniMatBench}~\citep{liu2026omnimatbenchhumancalibratedmultimodalreasoning},
MatSciBench~\citep{zhang2026matscibenchbenchmarkingreasoningability}, and
MaCBench~\citep{alampara2025probinglimitationsmultimodallanguage}.
OmniMatBench contains 3,171 expert-curated questions spanning 19 materials-science subfields
and evaluates both calculation-intensive reasoning (CAL) and open-ended question
answering (QA). MatSciBench consists of 1,340 materials reasoning problems, with
a particular emphasis on numerical calculation and formula-related reasoning.
MaCBench contains 1,153 multimodal questions covering more than 26
materials-related topics and a diverse range of scientific visual content.
Together, these benchmarks provide a comprehensive evaluation across the capability dimensions described above.

\textbf{Summary of Core Evaluation Results}
\begin{itemize}
    \item \textbf{Scientific Reasoning (MatSciBench).}
    \method{} achieves 63.59\% \RuleAcc{} on MatSciBench (+5.14 over its base
    model Intern-S2-Preview, 58.45\%), outperforming GPT-5.2 (59.6\%) and the
    same-scale open-weight Qwen3-VL-30B (58.06\%), while remaining competitive
    with much larger frontier proprietary models (Gemini~3.1~Pro, 74.01\%;
    Claude~Opus~4.6, 65.82\%).

    \item \textbf{Multimodal Understanding (MaCBench).}
    \method{} reaches 67.22\% \OverAcc{} (+2.17 over the base model), with the
    largest gains on structured visual tasks such as hand-drawn molecule
    recognition (+20.7), laboratory equipment (+20.0), and table question
    answering (+10.4).

    \item \textbf{Quantitative Calculation (OmniMatBench CAL).}
    \method{} improves \ESA{} from 21.98\% to 24.45\% (+2.47) while
    maintaining a 97.43\% completion rate across all 19 materials-science
    subfields.

    \item \textbf{Knowledge QA (OmniMatBench QA).}
    \MacroFOne{} remains comparable to the base model (27.59\% vs.\ 28.67\%),
    suggesting that calculation-oriented post-training leaves general
    question answering largely unchanged.

\end{itemize}
Figure~\ref{fig:overall_ranking} summarizes the comparisons against
representative foundation models.

\textbf{Main Contributions.} Our contributions are summarized as follows:
\begin{itemize}
    \item We develop \method{}, a unified materials foundation model built upon
    Intern-S2-Preview (35B-A3B), which provides a lightweight common backbone
    for materials knowledge, quantitative reasoning, and multimodal scientific
    understanding.

    \item We construct an approximately 150K materials-domain post-training
    corpus that combines teacher-guided reasoning distillation, scientific
    problem synthesis, and multi-rollout consistency filtering for
    high-quality reasoning supervision.

    \item We propose a retrieval-based multimodal data construction pipeline
    that combines text-to-image and image-to-image semantic retrieval over
    large scientific image pools to build task-specific multimodal training
    examples.

    \item We adopt a progressive training recipe that validates the data and
    post-training strategy on a 4B pilot model before scaling to the final 35B
    backbone, trained on an in-house Huawei Ascend 910B computing cluster.

    \item We conduct a comprehensive evaluation on three benchmarks, including
    our human-calibrated \textbf{OmniMatBench}, showing consistent improvements
    over the base model.
\end{itemize}
\section{\method{}: Model Overview and Capabilities}

\subsection{Model Architecture}

\method{} is built upon Intern-S2-Preview-35B, an efficient scientific multimodal foundation model with a 35B-A3B Mixture-of-Experts architecture. Intern-S2-Preview-35B is continually pretrained from Qwen3.5~\citep{qwen3.5} and retains the Qwen3.5-35B-A3B language backbone, while extending its capabilities to scientific multimodal understanding, reasoning, and agentic tasks through science-oriented pre-training and post-training.

Intern-S2-Preview provides a suitable starting point for materials specialization. On the one hand, it preserves the general-purpose language, reasoning, and multimodal capabilities of a modern foundation model. On the other hand, its science-oriented training exposes the model to heterogeneous scientific documents, figures, structured scientific content, and professional scientific tasks.

On top of this foundation, \method{} keeps the underlying multimodal architecture unchanged and specializes the model for materials science through domain-specific post-training. The post-training introduces materials-oriented supervision that spans domain knowledge, quantitative reasoning, structural understanding, and scientific visual interpretation. \method{} thereby inherits the general and scientific capabilities of Intern-S2-Preview while further strengthening its competence on materials-specific tasks.

\subsection{Model Development}
\begin{figure}[t]
    \centering \includegraphics[width=\linewidth]{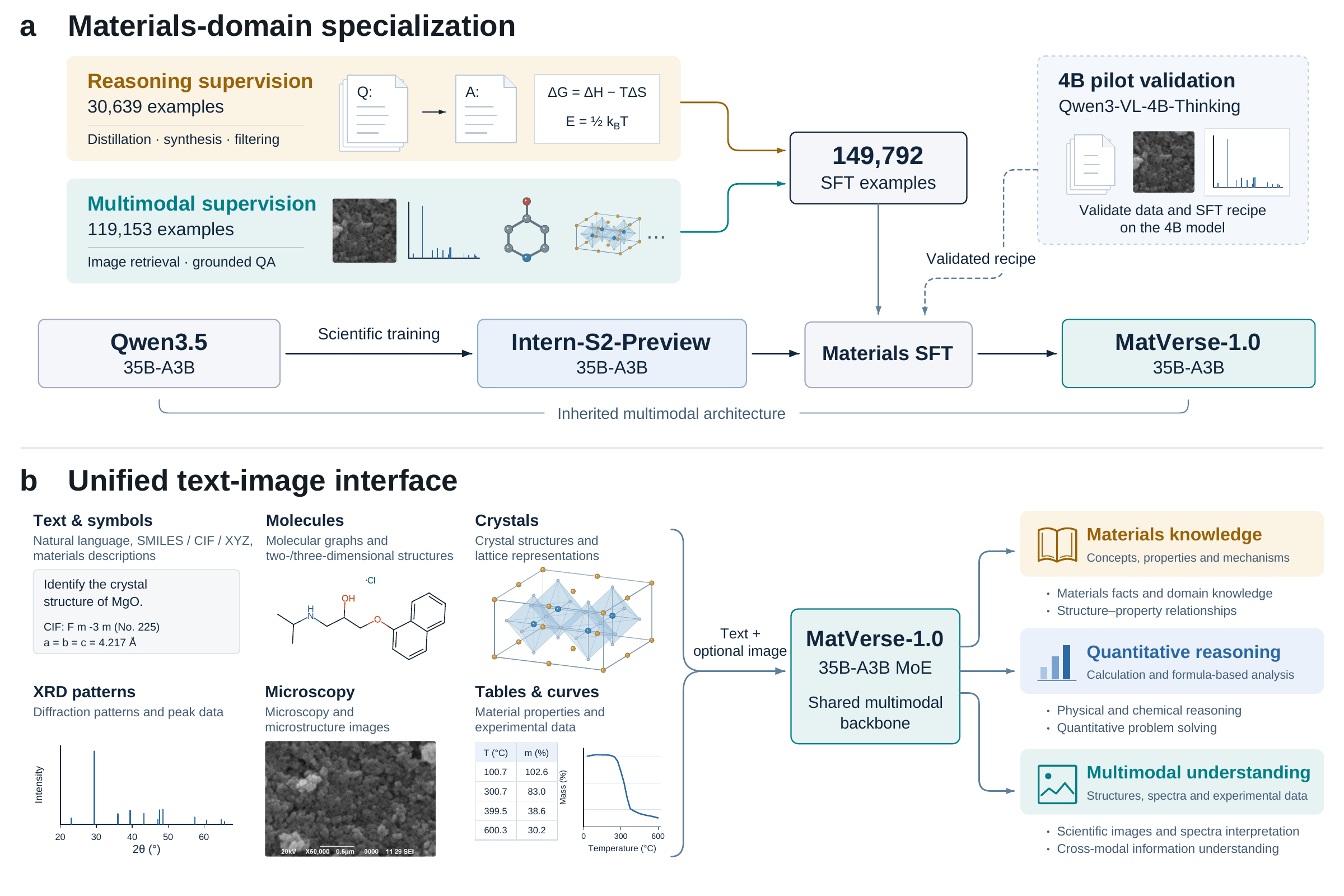}
    \caption{Overview of \method{}. (a) Materials-domain specialization:
    reasoning-oriented (30,639 examples) and multimodal (119,153 examples)
    supervision streams are merged into 149,792 SFT examples. The data and
    SFT recipe are validated on a Qwen3-VL-4B-Thinking pilot model before
    scaling, and the final model is obtained by materials-domain SFT of
    Intern-S2-Preview-35B-A3B, which inherits its multimodal architecture
    from the Qwen3.5 35B-A3B language backbone. (b) Unified text--image
    interface: \method{} accepts natural language and symbolic
    representations together with optional scientific imagery, and supports
    diverse materials-science tasks within a shared backbone.}
    \label{fig:model_pipeline}
\end{figure}
The development of \method{} (Figure~\ref{fig:model_pipeline}a) follows a progressive domain-specialization strategy: we first construct a large-scale materials corpus and validate the training recipe on a small pilot model, and then scale the validated pipeline to the final 35B-A3B model.

\textbf{Data collection.}
We collect materials-science data from multiple sources, including textbook-derived problem sets, online scientific resources, expert-curated question--answer pairs, and model-generated supervision. The resulting corpus contains both multimodal scientific data with associated images and text-only reasoning data. For text-oriented tasks, we further construct question--answer pairs and reasoning traces through model distillation~\citep{toshniwal2024openmathinstruct2acceleratingaimath}, followed by answer verification and consistency-based filtering. For multimodal tasks, existing materials problems and scientific descriptions are used as semantic anchors to retrieve relevant scientific images from large candidate pools, after which the retrieved image--text pairs are transformed into task-oriented multimodal training examples. In this way, the collected materials corpus is converted into unified supervision covering both domain knowledge and scientific multimodal understanding, yielding 149,792 SFT examples in total (Section~\ref{sec:data}).

\textbf{Pilot validation.}
Before large-scale training, we conduct pilot experiments on a Qwen3-VL-4B-Thinking model to evaluate the effectiveness of the constructed data and the domain-specific post-training recipe. We focus on supervised fine-tuning (SFT), with particular attention to instruction following, materials-domain reasoning, and the effective utilization of domain-specific knowledge. The pilot experiments show that SFT alone already provides substantial improvements on materials-science tasks (Section~\ref{sec:training}), indicating that targeted domain supervision can effectively adapt a strong scientific foundation model to materials-oriented problem settings.

\textbf{Scaling.}
We therefore adopt SFT as the primary post-training strategy and scale the validated recipe to the 35B-A3B Intern-S2-Preview backbone on an expanded materials-specific corpus, yielding the final \method{} model. The detailed data construction pipeline is introduced in Section~\ref{sec:data}, and the training configuration is described in Section~\ref{sec:training}.

\subsection{Capabilities}

\method{} is designed to support a broad range of materials-science tasks spanning domain knowledge, scientific reasoning, and multimodal scientific understanding. Rather than focusing on a single material system or task format, the model aims to provide a common foundation for interpreting and reasoning over the heterogeneous information encountered throughout materials research.

\textbf{Materials Knowledge and Scientific Reasoning.}
\method{} covers materials-science knowledge across fundamental concepts, atomic bonding and crystal structures, processing and manufacturing, and material degradation and failure, as well as the mechanical, thermal, electrical, magnetic, and optical properties of materials. It is further trained to solve quantitative and reasoning-intensive scientific problems, including numerical calculation, formula-based analysis, interpretation of structure--property relationships, and multi-step reasoning over experimental conditions. This coverage enables both knowledge-intensive question answering and calculation-oriented scientific reasoning.

\textbf{Multimodal Scientific Understanding.}
Beyond conventional text-based question answering, \method{} is designed to interpret the scientific representations commonly encountered in materials and molecular research. Representative inputs include molecular strings, two-dimensional molecular structures, multi-view three-dimensional atomic structures, crystallographic information, characterization spectra, microscopy images, scientific tables, and process or property curves. For molecular and structural understanding, the model can reason from symbolic molecular representations as well as two- and three-dimensional structural information: it can infer molecular properties from SMILES representations, identify local coordination geometries from atomic coordinates and structural renderings, and interpret crystallographic information from CIF files together with crystal-cell visualizations. It also supports the interpretation of experimental characterization data. Given X-ray diffraction patterns together with peak positions and intensities, \method{} can associate characteristic diffraction features with candidate crystal phases. It can further analyze microscopy images, scientific plots, tables, and experimental curves by extracting relevant features and combining them with domain knowledge. These capabilities allow \method{} to process materials information across multiple levels of scientific representation, from molecular and crystal structures to characterization results, processing conditions, and material properties.

\textbf{Towards AI-Assisted Materials Research.}
Together, the capabilities described above make \method{} suitable as a general-purpose backbone for AI-assisted materials research. It supports scientific question answering, interpretation of characterization results, structure and property analysis, and preliminary reasoning over experimental observations. More broadly, the unified text-and-image interface provides a natural basis for integration with automated research agents~\citep{qi2024large}. \method{} could serve as a reasoning and interaction layer for agentic scientific workflows when combined with external materials databases, simulation software, scientific computing tools, and laboratory automation systems. Within such workflows, it supports data retrieval, tool invocation, interpretation of computational and experimental results, and iterative decision making throughout the research process.

\section{Materials Data Construction}
\label{sec:data}

\begin{figure}[t]
    \centering
    \includegraphics[width=\linewidth]{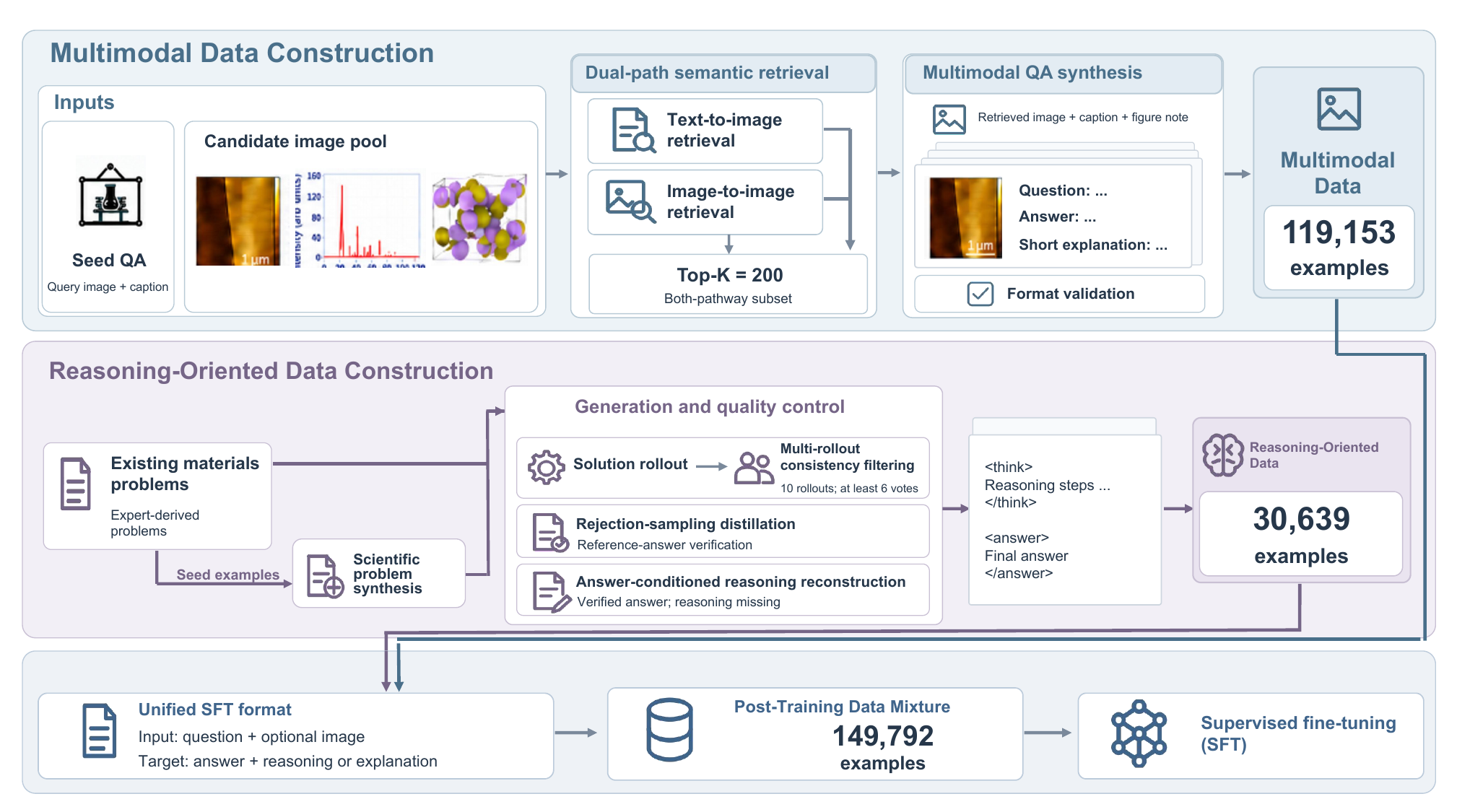}
    \caption{Materials data construction for MatVerse. The reasoning-oriented stream synthesizes new problems from seed examples and applies teacher-guided solution generation and quality control according to answer availability. The multimodal stream uses text-to-image and image-to-image retrieval to select scientific images for generating image-grounded questions, answers, and short explanations. The two streams contribute 30,639 and 119,153 examples, respectively, yielding 149,792 post-training examples in total.}
    \label{fig:materials-data-construction}
\end{figure}

To further strengthen the base model in materials knowledge, scientific reasoning, and multimodal scientific understanding, we construct a large-scale post-training corpus tailored to materials science. Since different materials tasks require substantially different forms of supervision, our data construction framework combines task-specific data generation with a shared set of teacher-guided refinement and quality-control procedures.

At a high level (Figure~\ref{fig:materials-data-construction}), the corpus
consists of two complementary data streams. The
first is a \emph{reasoning-oriented stream}, which focuses on materials
knowledge, quantitative calculation, and scientific problem solving. The
second is a \emph{multimodal stream}, which covers scientific images,
structural representations, experimental characterization, tables, and other
visually grounded materials data. The two streams are constructed with
different data generation strategies: reasoning data are primarily developed
through rejection-sampling distillation and consistency-based filtering,
whereas multimodal data are constructed through image-grounded
question--answer synthesis. Strong teacher models are employed throughout
these processes to generate high-quality supervision and, where applicable,
to support reasoning reconstruction and consistency filtering. All examples
are finally converted into a unified supervised fine-tuning format, resulting
in approximately 150K materials-domain post-training samples.

\subsection{Teacher-Guided Data Generation and Quality Control}
\label{sec:teacher_guided_data}

A central component of our data construction pipeline is the use of strong teacher models to generate, refine, and validate training examples. Rather than relying on a single distillation strategy, we employ different teacher-guided procedures depending on whether reliable reference answers are available and whether the objective is solution verification, reasoning enrichment, or synthetic-data filtering.

\textbf{Rejection-sampling distillation.}
For samples with reference answers, the teacher model is prompted to solve the problem independently without access to the reference answer during generation. We extract the predicted final answer from each rollout and compare it with the reference answer using task-specific rule-based evaluators. Only trajectories whose final answers pass the corresponding verification are retained as supervised examples. Incorrect generations are resampled until a valid trajectory is obtained or a predefined attempt limit is reached.

We use multiple strong teacher models in the rejection-sampling pipeline, including Qwen3.7-Max, Qwen3-235B, Claude Opus 4.8 Thinking, and Qwen3.6-Max-Preview. They are applied to different materials reasoning tasks to provide diverse teacher-generated solution trajectories.

\textbf{Answer-conditioned reasoning reconstruction.}
For some training examples, the final answer is already known and considered reliable, while an explicit reasoning trajectory is unavailable. In this setting, we provide the verified final answer to the teacher model and ask it to reconstruct a coherent step-by-step solution leading to that answer. Optional reference key points may also be provided as hidden guidance when available.

Unlike rejection-sampling distillation, this procedure is explicitly
answer-conditioned: rather than evaluating whether the teacher can solve a
problem independently, it enriches an already verified example with a
structured intermediate reasoning process. We apply this strategy to textual
examples with verified answers. For multimodal
examples, questions, answers, and short explanations are instead generated
jointly during the initial synthesis process, as described in
Section~\ref{sec:multimodal_data}.

\textbf{Multi-rollout consistency filtering.}
For synthesized examples whose correctness cannot be guaranteed directly by an existing reference answer, we further use agreement across multiple independent teacher rollouts as a quality signal. For each generated problem, the teacher model independently produces ten candidate solutions. Their final answers are normalized and grouped according to answer equivalence, and a sample is retained only when the most frequent answer receives at least six votes out of ten.

Among the rollouts associated with the majority answer, we preferentially select one containing a complete and non-empty reasoning trajectory. This process filters out ambiguous or unstable synthetic examples before they are added to the final post-training corpus.

\subsection{Reasoning-Oriented Data Construction}
\label{sec:reasoning_data}

The reasoning-oriented portion of the corpus is designed to strengthen materials-domain knowledge, quantitative calculation, and scientific problem solving. It combines expert-derived problems with additional synthetic examples generated from existing seed questions. The teacher-guided distillation and quality-control procedures described in Section~\ref{sec:teacher_guided_data} are applied where appropriate to enrich and filter these examples.

\textbf{Scientific problem synthesis.}
The number of expert-authored materials problems is limited compared with the scale required for post-training. We therefore expand the reasoning corpus by synthesizing new scientific problems from existing seed examples. The synthesis process preserves the underlying scientific topic, problem type, reasoning pattern, and expected answer structure, while modifying problem-specific details when scientifically appropriate.

For numerical problems, the augmentation primarily changes numerical values, materials, units, and physical or experimental conditions while preserving the original reasoning path. For formula-oriented problems, the model may modify symbolic variables, physical constants, material contexts, or crystallographic directions and planes while maintaining the underlying scientific relation.

Generated questions are required to remain physically meaningful, solvable, and consistent with established materials-science principles. The generated problems are subsequently solved and filtered using these teacher-guided procedures. The complete synthesis and rollout prompts are provided in Appendix~\ref{app:prompts}. In total, this stream contributes 30,639 reasoning-oriented examples.

\subsection{Multimodal Data Construction}
\label{sec:multimodal_data}

Materials-science visual data differ substantially from general natural-image
data and often encode highly specialized scientific information, including
diffraction patterns, microscopy images, molecular and crystal structures,
scientific tables, patent figures, and experimental equipment. To expand
multimodal supervision while retaining realistic scientific content, we use
MaCBench images as queries to retrieve related images from a large candidate
pool and then synthesize new questions directly on the retrieved images.
Existing captions and figure notes are used as grounding context during
synthesis rather than being reused as training questions.

\textbf{Dual-path semantic retrieval.}
Each MaCBench image is used as a query in two complementary retrieval
pathways, text-to-image and image-to-image, with the top-$K$ candidates
retained from each pathway ($K=200$). In the \emph{text-to-image} pathway,
the caption associated with the query image is embedded and matched against
candidate images, emphasizing semantic correspondence. In the
\emph{image-to-image} pathway, the query image itself is embedded and
compared with candidate images, placing greater weight on visual structure
and particularly benefiting scientific data with characteristic visual
patterns, such as diffraction and microscopy images.

The two pathways provide complementary retrieval results, but we do not use
their union for subsequent synthesis. Instead, we retain only records that
are returned by both pathways and remain linked to a MaCBench query. This
intersection contains 52,922 records, corresponding to 52,902 unique
retrieved images. We further filter these records by a retrieval-score
threshold of $0.60$, resulting in 52,246 retained records.

\textbf{Multimodal QA synthesis.}
For each retained record, a multimodal teacher generates new question--answer
pairs grounded in the retrieved image. The linked MaCBench example provides
the task type and answer format, either multiple-choice or open-ended, but
its original question text is not reused. Instead, the retrieved image,
its existing caption, and the associated figure note are provided as
grounding context, and the teacher is instructed to formulate new questions
about the image. A post-hoc BLEU-based check against the original MaCBench
wording is used to identify potentially overlapping questions; flagged
examples are retained with an annotation rather than deleted.

The teacher generates the question, answer, and a short explanation jointly,
yielding 119,167 synthesized pairs. After format validation, we retain
119,153 examples (111,598 multiple-choice and 7,555 open-ended). The resulting
multimodal corpus covers a broad range of scientific visual content,
including molecular and crystal structures, diffraction patterns, microscopy,
laboratory equipment, tables, and scientific figures. Unlike the
textual reasoning data, these multimodal examples are not subjected to a
separate answer-conditioned reasoning rewrite, since their explanations are
generated jointly with the questions and answers during synthesis.

\subsection{Post-Training Data Mixture}
\label{sec:data_mixture}

After synthesis, distillation, retrieval, reasoning reconstruction, and quality filtering, all examples are converted into a unified supervised fine-tuning format. The final post-training corpus contains 149,792 samples. Table~\ref{tab:data_mixture} summarizes the overall data composition.

\begin{table}[t]
\centering
\caption{Composition of the materials post-training corpus.}
\label{tab:data_mixture}
\small
\begin{tabular}{lrr}
\toprule
\textbf{Data category} & \textbf{\# Samples} & \textbf{Ratio} \\
\midrule
Broad materials reasoning and synthesis & 27,101 & 18.1\% \\
Quantitative materials reasoning         & 3,538  & 2.4\%  \\
Multimodal multiple-choice data          & 111,598 & 74.5\% \\
Multimodal open-ended data               & 7,555   & 5.0\%  \\
\midrule
\textbf{Total}                           & \textbf{149,792} & \textbf{100\%} \\
\bottomrule
\end{tabular}
\end{table}

The resulting mixture is dominated by multimodal scientific supervision while retaining high-quality materials reasoning data. Together, these complementary data sources cover materials-domain knowledge, quantitative and scientific reasoning, structural and experimental understanding, and multimodal scientific question answering.

 \section{Training}
\label{sec:training}

\subsection{Progressive Training Strategy}

The training of \method{} follows a progressive scaling strategy from
small-scale validation to large-scale specialization. We first validate the
data construction pipeline and the materials-specific post-training recipe on
Qwen3-VL-4B-Thinking. Following the empirical scaling intuition that model
capacity, training data, and computation should be scaled
jointly~\citep{kaplan2020scalinglawsneurallanguage}, we then scale the validated
recipe to the 35B-A3B Intern-S2-Preview backbone, resulting in the final
\method{} model.

Throughout this process, supervised fine-tuning (SFT) serves as the primary
post-training objective. The materials-domain data constructed in
Section~\ref{sec:data} are converted into a unified supervised training format.
The inputs include both text-only scientific problems and multimodal content
involving scientific images, tables, and structural representations. The
target outputs contain direct answers, explicit scientific reasoning
trajectories, or both, depending on the task. Under this unified objective,
all capability dimensions are reinforced jointly within a single training
process.

\subsection{Training Recipe Validation on the 4B Model}

Before large-scale training, we use Qwen3-VL-4B-Thinking as the pilot model to
examine whether the constructed materials-domain supervision can effectively
improve domain-specific capabilities and to establish the basic post-training
recipe used in subsequent scaling.

The 4B stage includes both component-level experiments and full-mixture
training. In the component-level experiments, reasoning-oriented supervision
and multimodal materials data are trained separately to examine the effects of
different types of domain supervision. We then combine the different data
sources and perform full-mixture materials-specific fine-tuning to validate the
overall data construction and training pipeline.

Training is performed with Megatron-based distributed SFT through the
ms-swift training stack on 256 Ascend NPUs. We apply full-parameter
fine-tuning to the language model components while keeping the pretrained
visual encoder and multimodal alignment modules frozen. Sequence packing is
applied to improve computational efficiency for training examples with
heterogeneous sequence lengths, and the maximum sequence length is set
sufficiently long to accommodate scientific reasoning trajectories and complex
multimodal contexts. The complete hyperparameter settings are provided in
Appendix~\ref{app:hyperparameters}.

The pilot experiments show that materials-specific supervision consistently
improves performance across multiple materials-science tasks: on the CAL
track of OmniMatBench, \ESA{} improves from 8.12\% for the base
Qwen3-VL-4B-Thinking to 20.78\% with the full materials data mixture
(Figure~\ref{fig:pilot_scaling}a). We therefore use
the validated data recipe and SFT-based post-training paradigm as the basis for
subsequent large-scale specialization.

\subsection{Scaling to the 35B-A3B Model}

After validating the training strategy on the 4B model, we further scale
materials-specific post-training to Intern-S2-Preview-35B-A3B, which serves as
the backbone of the final \method{} model.

Compared with the 4B pilot model, Intern-S2-Preview provides substantially
greater model capacity and has already undergone science-oriented pre-training
and post-training, which offers a stronger foundation in scientific knowledge,
reasoning, and multimodal understanding. At this stage, we use the complete
materials-domain corpus described in Section~\ref{sec:data}, combining
materials knowledge and reasoning data, synthesized scientific problems, and
retrieval-constructed multimodal supervision.

For the 35B-A3B stage, we follow the model architecture and large-scale
training paradigm of Intern-S2-Preview~\citep{bai2026intern}, while transferring
the materials data construction and domain-supervision strategy validated
during the 4B experiments.

\subsection{Implementation Details}

Model training is conducted on an in-house AI computing cluster equipped with
Huawei Ascend 910B NPUs. The cluster supports large-scale distributed training
through high-bandwidth RDMA interconnects and provides elastic computing
resources for experiments at different model scales.

For the final 35B-A3B model, we use MindSpeed-MM as the large-scale
multimodal training framework and adopt Fully Sharded Data Parallel (FSDP) for
distributed training. FSDP shards model parameters, gradients, and optimizer
states across devices, enabling full-parameter post-training of the large model
under practical per-device memory constraints. Apart from the
materials-specific training data and supervision strategy, the underlying model
configuration follows the training paradigm of Intern-S2-Preview.

\section{Evaluation}
\label{sec:evaluation}

We evaluate \method{} on three complementary materials-science benchmarks:
OmniMatBench~\citep{liu2026omnimatbenchhumancalibratedmultimodalreasoning},
MatSciBench~\citep{zhang2026matscibenchbenchmarkingreasoningability}, and
MaCBench~\citep{alampara2025probinglimitationsmultimodallanguage}. Together,
these benchmarks cover three central capability dimensions for materials-science
foundation models: broad domain knowledge, scientific reasoning, and
multimodal understanding of scientific information. Their complementary task
formulations allow us to evaluate how materials-specific post-training improves
different aspects of scientific capability.

\subsection{Benchmarks}

\textbf{OmniMatBench}~\citep{liu2026omnimatbenchhumancalibratedmultimodalreasoning}.
OmniMatBench is a human-calibrated benchmark designed to evaluate broad
materials-science knowledge and reasoning across 19 subfields. It combines
knowledge-intensive question answering with calculation-oriented scientific
problems, providing a comprehensive assessment of materials understanding.
The benchmark contains 3,171 expert-curated questions spanning four broad areas:
fundamental materials science, structural and engineering materials, materials
processing and manufacturing, and functional and applied materials. It consists
of two complementary tracks: the CAL track with 1,910 calculation-oriented
problems, and the QA track with 1,261 open-ended questions. We report the two
tracks separately to characterize quantitative reasoning and materials knowledge
understanding.

\textbf{MatSciBench}~\citep{zhang2026matscibenchbenchmarkingreasoningability}.
MatSciBench focuses on scientific reasoning in materials science. It contains
1,340 college-level problems collected from core undergraduate and graduate
materials-science textbooks. The benchmark covers six major areas (materials,
properties, structures, fundamental mechanisms, processing, and failure
mechanisms) and includes 31 fine-grained subfields. Most problems require
quantitative reasoning, while additional subsets evaluate formula derivation
and symbolic reasoning. Moreover, 315 problems contain visual information,
enabling evaluation of both text-based and visually grounded scientific
reasoning. Difficulty and problem-type annotations are retained for detailed
capability analysis.

\textbf{MaCBench}~\citep{alampara2025probinglimitationsmultimodallanguage}.
MaCBench evaluates multimodal scientific understanding in materials and chemistry.
It contains 1,153 questions spanning more than 26 scientific topics and includes
diverse visual information encountered in materials research, such as scientific
tables, plots, diffraction patterns, microscopy images, molecular and crystal
representations, and experimental figures. The benchmark covers capabilities
ranging from visual information extraction to multimodal scientific reasoning.
Its capability annotations and task-family labels enable detailed analysis of
different aspects of materials-oriented multimodal understanding.

\subsection{Baselines}

We evaluate \method{} from two complementary perspectives: improvement over its
underlying foundation model and comparison with representative general-purpose
and scientific foundation models. First, we compare \method{} with its
corresponding base model Intern-S2-Preview to directly measure the effect of
materials-specific post-training. We further compare \method{} with strong general-purpose language and
multimodal models. These baselines span both open-weight models available on
Hugging Face (e.g., the Qwen, Intern, GLM, Gemma, and Llama series) and
proprietary frontier models accessed through official APIs (e.g., the GPT,
Claude, Gemini, and Grok series).

\subsection{Evaluation Metric}

We follow the official evaluation method of each benchmark and preserve its
task-specific scoring procedure. All models are evaluated under consistent
prompting and inference configurations. Unless otherwise specified, no external
retrieval system, scientific database, or tool augmentation is used during
inference.

\textbf{Primary metrics.}
We report one primary metric for each evaluation setting and use these names
consistently throughout the paper:
\begin{itemize}
    \item \textbf{\ESA{} (ESA)} for OmniMatBench CAL:
    model answers are parsed into structured answer slots and compared with
    reference slots by exact matching.
    
    \item \textbf{\MacroFOne{}} for OmniMatBench QA:
    open-ended answers are evaluated using the official rubric-based matching
    pipeline.
    
    \item \textbf{\RuleAcc{}} for MatSciBench:
    the official rule-based evaluator extracts final answers and checks them
    against task-specific answer rules.
    
    \item \textbf{\OverAcc{}} for MaCBench:
    the official task-aware evaluator assigns question-level correctness based
    on the corresponding task format.
\end{itemize}

\section{Main Results}
\label{sec:main_results}

We organize the main results around three questions. First, how does
\method{} compare with Intern-S2-Preview and with representative
general-purpose and scientific foundation models under the official
evaluation settings of each benchmark? Second, where do the improvements
concentrate? Third, what trade-offs and failure modes emerge from domain
specialization?

\subsection{Overall Comparison}

\begin{table}[H]
\centering
\caption{Overall primary-metric results on the three benchmarks.}
\label{tab:overall_main}
\small
\begin{tabular}{llccc}
\toprule
\textbf{Benchmark} & \textbf{Metric} & \textbf{Base / Ref.} &
\textbf{\method{}} & \textbf{$\Delta$} \\
\midrule
MatSciBench & \RuleAcc{} & 58.45\% & 63.59\% & +5.14 \\
MaCBench & \OverAcc{} & 65.05\% & 67.22\% & +2.17 \\
OmniMatBench (CAL) & ESA & 21.98\% & 24.45\% & +2.47 \\
OmniMatBench (QA) & \MacroFOne{} & 28.67\% & 27.59\% & -1.08 \\
\bottomrule
\end{tabular}
\end{table}

Table~\ref{tab:overall_main} summarizes the primary metrics on the three
benchmarks. Following materials-specific post-training, \method{} improves
over Intern-S2-Preview on MatSciBench, MaCBench, and the CAL track of
OmniMatBench, while the QA track remains essentially unchanged ($-$1.08).
Relative to other models, \method{} outperforms GPT-5.2 (59.6\%) and the
same-scale open-weight Qwen3-VL-30B (58.06\%) on MatSciBench, while larger
frontier systems such as Gemini~3.1~Pro (74.01\%) and Claude~Opus~4.6
(65.82\%) remain stronger. On MaCBench and
OmniMatBench CAL, \method{} clearly improves over its base model, while
frontier systems such as Gemini~3.1~Pro and Claude~Opus~4.7 remain ahead.

\begin{figure}[!t]
\centering
\includegraphics[width=0.80\linewidth]{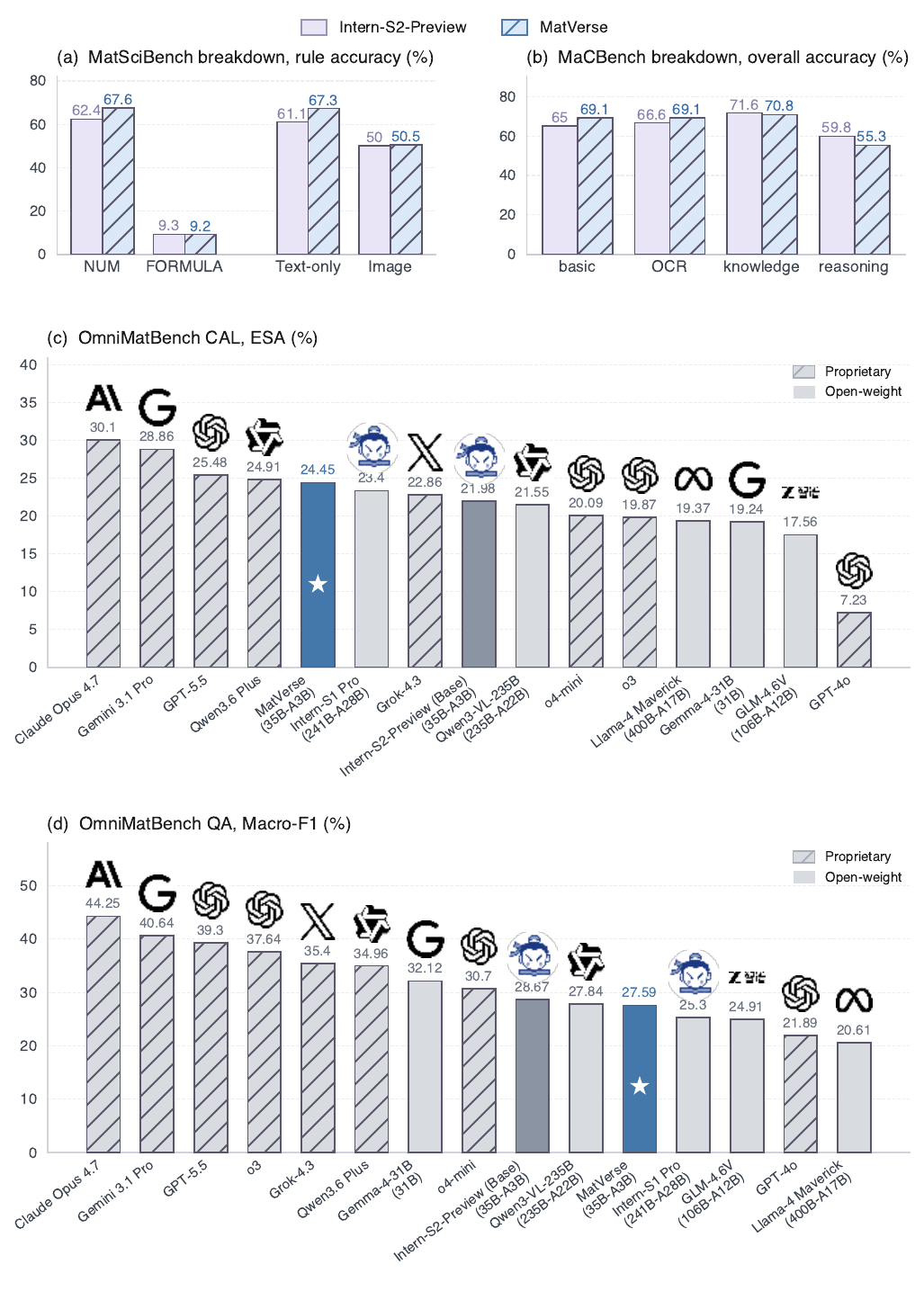}
\caption{Fine-grained results of \method{}
(Intern-S2-Preview in pale bars, \method{} in blue, hatched in a--b).
(a) MatSciBench breakdown by problem type and modality.
(b) MaCBench breakdown by capability tag.
(c) OmniMatBench CAL (\ESA{}) and (d) QA (\MacroFOne{})
compared with representative foundation models.
Models with diagonally hatched bars are proprietary systems accessed via
API; all other models, including \method{}, are open-weight and available
on Hugging Face.}
\label{fig:main_results}
\end{figure}

\subsection{Where the Gains Come From}

The improvements are concentrated in quantitative and structured scientific
reasoning (Figure~\ref{fig:main_results}a). On MatSciBench, \RuleAcc{}
increases from 62.4\% to 67.6\% on numerical problems and from 61.1\% to
67.3\% on text-only problems, whereas FORMULA problems remain essentially
unchanged (9.3\% vs.\ 9.2\%) and image-containing problems stay at around
50\% (50.0\% vs.\ 50.5\%).

\begin{figure}[!t]
\centering
\includegraphics[width=0.85\linewidth]{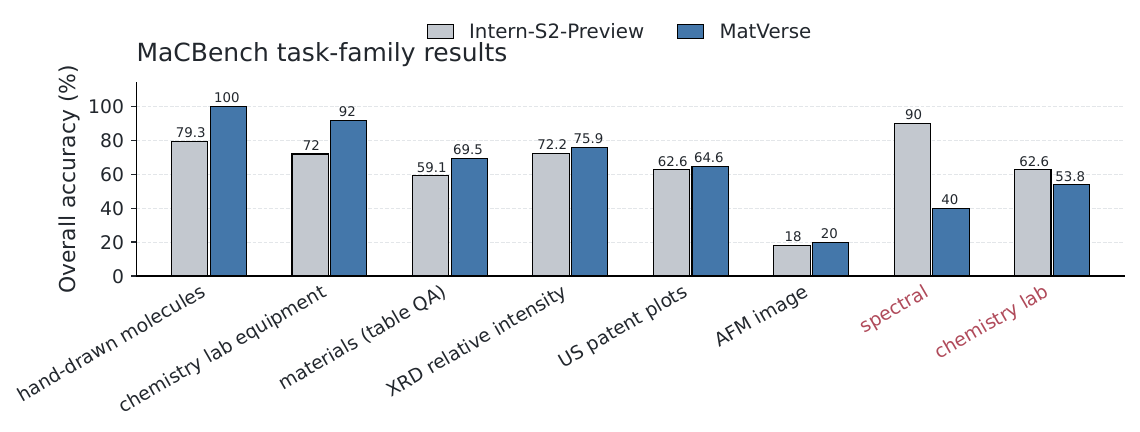}
\caption{MaCBench task-family results (overall accuracy, \%;
Intern-S2-Preview vs.\ \method{}). Regressed task families are highlighted
in red.}
\label{fig:analysis}
\end{figure}

On the multimodal side (Figure~\ref{fig:main_results}b), \method{} improves
basic and OCR-tagged questions, from 65.0\% to 69.1\% and from 66.6\% to
69.1\%, respectively. At the task-family level
(Figure~\ref{fig:analysis}), the largest gains are observed for
hand-drawn molecules (+20.7 percentage points), laboratory equipment
(+20.0), and table question answering (+10.4), suggesting that
materials-specific multimodal post-training is particularly effective for
connecting visual information with structured scientific concepts, especially
in tasks involving scientific diagrams, laboratory objects, and tabular
information.

\subsection{Trade-offs and Failure Modes}

Domain specialization also introduces trade-offs. On OmniMatBench QA,
\method{} scores 27.59\% \MacroFOne{}, 1.08 points below its base model
(28.67\%), while the strongest general-purpose models remain considerably
ahead (Claude~Opus~4.7 at 44.25\%, Gemini~3.1~Pro at 40.64\%, and GPT-5.5 at
39.30\%). A similar pattern appears on MaCBench reasoning, which decreases
from 59.8\% to 55.3\%, and on individual task families such as spectral
(90.0\% to 40.0\%) and chemistry\_lab (62.6\% to 53.8\%). Capabilities that
were already weak before post-training show little movement: FORMULA problems
on MatSciBench stay near 9\% \RuleAcc{}, image-containing problems near 50\%,
and AFM recognition near 20\% \OverAcc{}.

Across OmniMatBench subfields (Table~\ref{tab:omnimat_cal_subfields}), CAL is
highest on gem materials (42.00\%), powder materials (41.84\%), and materials
chemistry (41.14\%), and lowest on inorganic non-metallic materials (0.00\%)
and metallurgical engineering (1.30\%). QA does not follow the same order:
welding technology reaches 54.04\%, materials chemistry 48.29\%, and
inorganic non-metallic materials 41.00\% \MacroFOne{} despite a CAL score of
0.00\%. These results show that \method{} exhibits heterogeneous performance
across materials subfields, reflecting the varying difficulty and knowledge
requirements of different domains.

\begin{table}[ht]
\centering
\caption{\method{} on OmniMatBench by subfield.
CAL reports \ESA{} and the completion rate; QA reports \MacroFOne{}.}
\label{tab:omnimat_cal_subfields}
\footnotesize
\setlength{\tabcolsep}{4pt}
\begin{tabular}{clrrrrr}
\toprule
\textbf{ID} & \textbf{Subfield} & \textbf{CAL \#Q} & \textbf{ESA} & \textbf{Compl.} & \textbf{QA \#Q} & \textbf{\MacroFOne{}} \\
\midrule
01 & Materials Science and Engineering & 78 & 36.32\% & 100\% & 85 & 12.10\% \\
02 & Soft Matter Science and Engineering & 64 & 17.26\% & 98.44\% & 70 & 33.69\% \\
03 & Materials Physics & 91 & 30.00\% & 97.80\% & 75 & 28.99\% \\
04 & Materials Chemistry & 96 & 41.14\% & 97.92\% & 80 & 48.29\% \\
05 & Materials Design Science and Engineering & 87 & 24.17\% & 98.85\% & 78 & 36.77\% \\
06 & Metallurgical Engineering & 86 & 1.30\% & 98.84\% & 96 & 12.89\% \\
07 & Welding Technology and Engineering & 87 & 28.14\% & 96.55\% & 84 & 54.04\% \\
08 & Intelligent Materials and Structures & 168 & 10.72\% & 98.21\% & 36 & 12.78\% \\
09 & New Energy Materials and Devices & 166 & 27.56\% & 98.80\% & 16 & 8.91\% \\
10 & Metallic Materials & 95 & 30.41\% & 96.84\% & 86 & 38.99\% \\
11 & Inorganic Non metallic Materials & 87 & 0.00\% & 100\% & 44 & 41.00\% \\
12 & Polymer Materials & 99 & 23.74\% & 96.97\% & 62 & 38.67\% \\
13 & Composite Materials & 88 & 32.58\% & 97.73\% & 80 & 40.36\% \\
14 & Powder Materials & 101 & 41.84\% & 93.07\% & 81 & 32.64\% \\
15 & Gem Materials and Gemology & 75 & 42.00\% & 100\% & 87 & 28.05\% \\
16 & Functional Materials & 131 & 15.46\% & 98.47\% & 35 & 18.77\% \\
17 & Electronic Information Materials & 140 & 16.33\% & 97.14\% & 91 & 13.90\% \\
18 & Nanomaterials and Technology & 87 & 31.69\% & 94.25\% & 11 & 7.86\% \\
19 & Composite Materials Molding & 84 & 13.80\% & 90.48\% & 64 & 15.59\% \\
\midrule
ALL & 19 categories & 1,910 & 24.45\% & 97.43\% & 1,261 & 27.59\% \\
\bottomrule
\end{tabular}
\end{table}

\FloatBarrier
\section{Analysis}
\label{sec:analysis}

We further analyze the evaluation results from three perspectives:
scientific reasoning, multimodal scientific understanding, and domain-specific
capability coverage.

\textbf{Enhanced scientific reasoning.} MatSciBench results demonstrate that materials-specific post-training
effectively improves scientific reasoning capabilities. The largest improvements
are observed on structured numerical problems, indicating that domain-specific
supervision helps the model better organize and utilize scientific knowledge for
multi-step quantitative reasoning. These results highlight the importance of
curating materials-oriented scientific problems for improving reasoning ability
in specialized domains.

\textbf{Multimodal understanding.} Results on MaCBench show that \method{} improves the interpretation of diverse
scientific visual information. In particular, gains on structured visual content
such as tables, molecular illustrations, and laboratory-related images suggest
that materials-oriented multimodal supervision strengthens the connection
between visual evidence and scientific concepts. This capability is important
for practical materials workflows, where experimental information is often
represented in heterogeneous visual formats.

\textbf{Domain coverage.} OmniMatBench results demonstrate that \method{} can generalize across a wide
range of materials subfields. Performance varies across different scientific
areas, reflecting the inherent diversity and specialization of materials
science. These results suggest that continued expansion of domain-specific data
can further improve scientific coverage and support more comprehensive
materials research applications.

\begin{figure}[!ht]
\centering
\includegraphics[width=\linewidth]{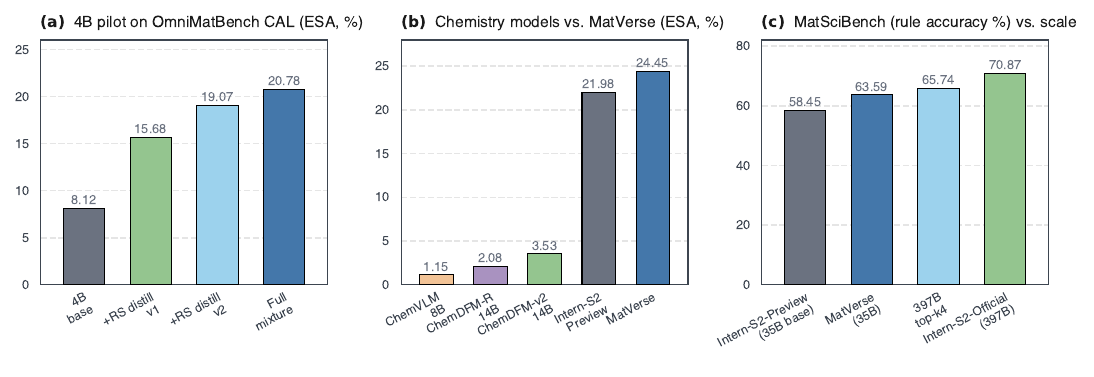}
\caption{Data recipe, domain gap, and scale analysis.
(a)~OmniMatBench CAL \ESA{} of the Qwen3-VL-4B-Thinking pilot as the data
recipe progresses from the base model through rejection-sampling (RS)
distillation data v1/v2 to the full materials mixture.
(b)~\ESA{} of chemistry-specific models, the 35B base model, and \method{}
on OmniMatBench CAL.
(c)~MatSciBench \RuleAcc{} across model scales, including the 397B
Intern-S2-Official baseline.}
\label{fig:pilot_scaling}
\end{figure}

\textbf{Data recipe, domain gap, and scale.}
Figure~\ref{fig:pilot_scaling} summarizes three further perspectives.
First, the 4B pilot ablation (Figure~\ref{fig:pilot_scaling}a) shows that
each stage of the data pipeline---rejection-sampling distillation and the
final full mixture---contributes consistent gains on OmniMatBench CAL,
improving \ESA{} from 8.12\% to 20.78\% ($2.6\times$) before scaling.
Second, chemistry-specific models such as ChemVLM and ChemDFM obtain at
most 3.53\% \ESA{} on the CAL track (Figure~\ref{fig:pilot_scaling}b),
roughly an order of magnitude below \method{} (24.45\%), indicating that
chemistry-oriented supervision alone does not transfer to
calculation-oriented materials problems.
Third, on MatSciBench, \method{} (63.59\% \RuleAcc{}) sits between its 35B
base model (58.45\%) and the much larger Intern-S2-Official 397B baseline
(70.87\%) (Figure~\ref{fig:pilot_scaling}c), suggesting that
domain-specific post-training and model scale provide complementary
pathways to stronger materials reasoning.

\textbf{Towards research assistance.} Beyond standalone benchmark evaluation, the capabilities developed by
\method{} provide a foundation for integrating materials models with external
scientific tools, databases, and automated research systems. Combining domain
knowledge, scientific reasoning, and multimodal understanding enables potential
applications in materials discovery, experiment analysis, and AI-assisted
research workflows.

\begin{figure}[!ht]
\centering
\begin{minipage}[t]{0.48\linewidth}
  \centering
  \includegraphics[width=0.62\linewidth]{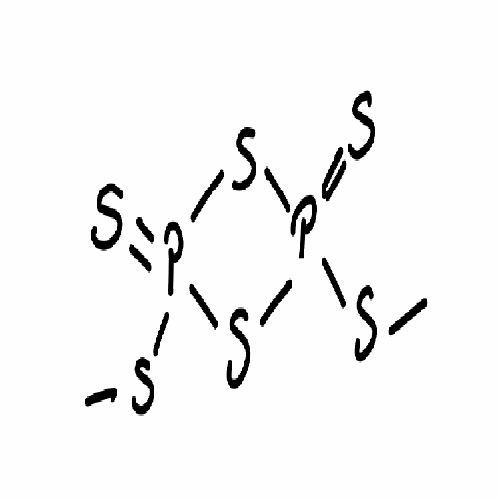}\par\smallskip
  {\footnotesize\raggedright
  \textbf{(a) Q:} What is a valid SMILES representation for the hand-drawn
  molecule?\par
  \textbf{MatVerse:} Cyclic molecule with two P atoms and three S atoms;
  each P is double-bonded to S and single-bonded to a methylthio group;
  the valid SMILES is \texttt{CSP1(=S)SP(=S)(S1)SC} \textbf{(A)}.\par}
\end{minipage}\hfill
\begin{minipage}[t]{0.48\linewidth}
  \centering
  \includegraphics[width=0.62\linewidth]{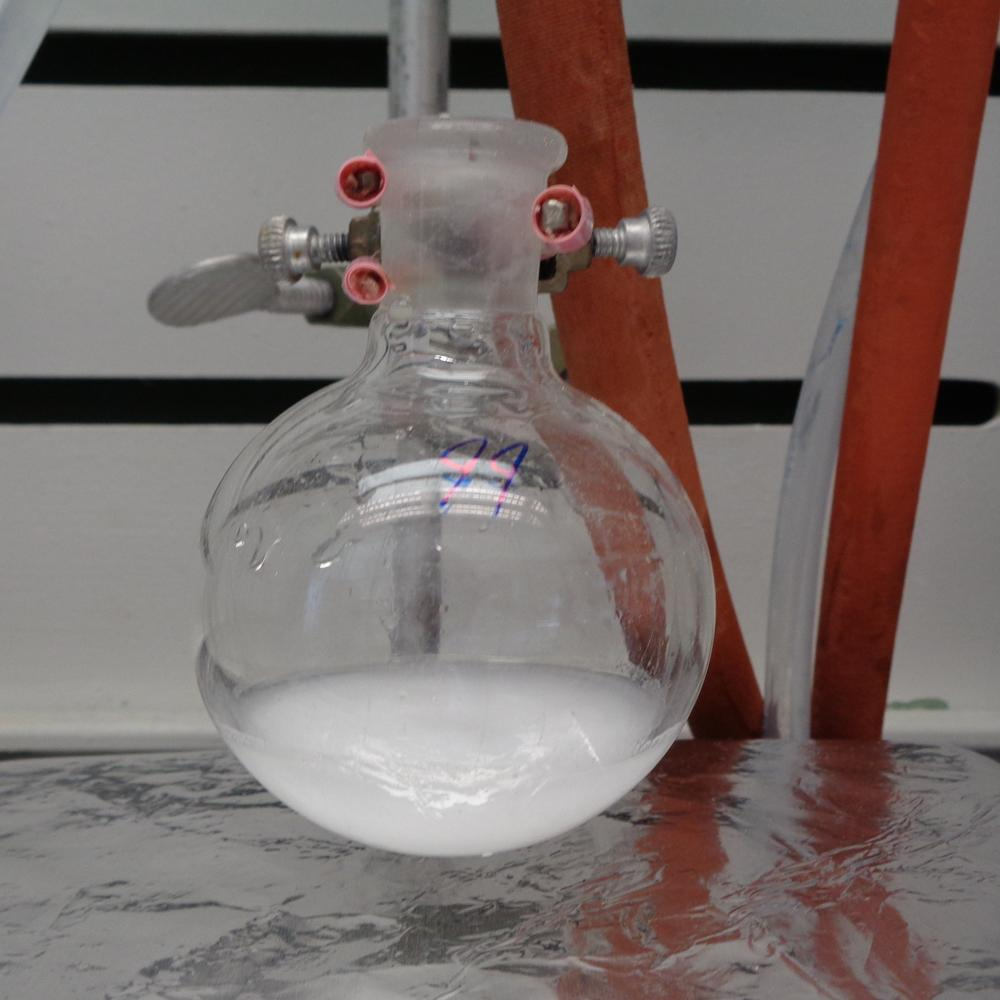}\par\smallskip
  {\footnotesize\raggedright
  \textbf{(b) Q:} Which chemistry-lab equipment do you see?\par
  \textbf{MatVerse:} Spherical body with a narrow neck, used for heating or
  mixing; unlike beakers, funnels, or columns $\to$ \textbf{Round Flask}
  (B).\par}
\end{minipage}
\par\medskip
\begin{minipage}[t]{0.48\linewidth}
  \centering
  \includegraphics[width=0.62\linewidth]{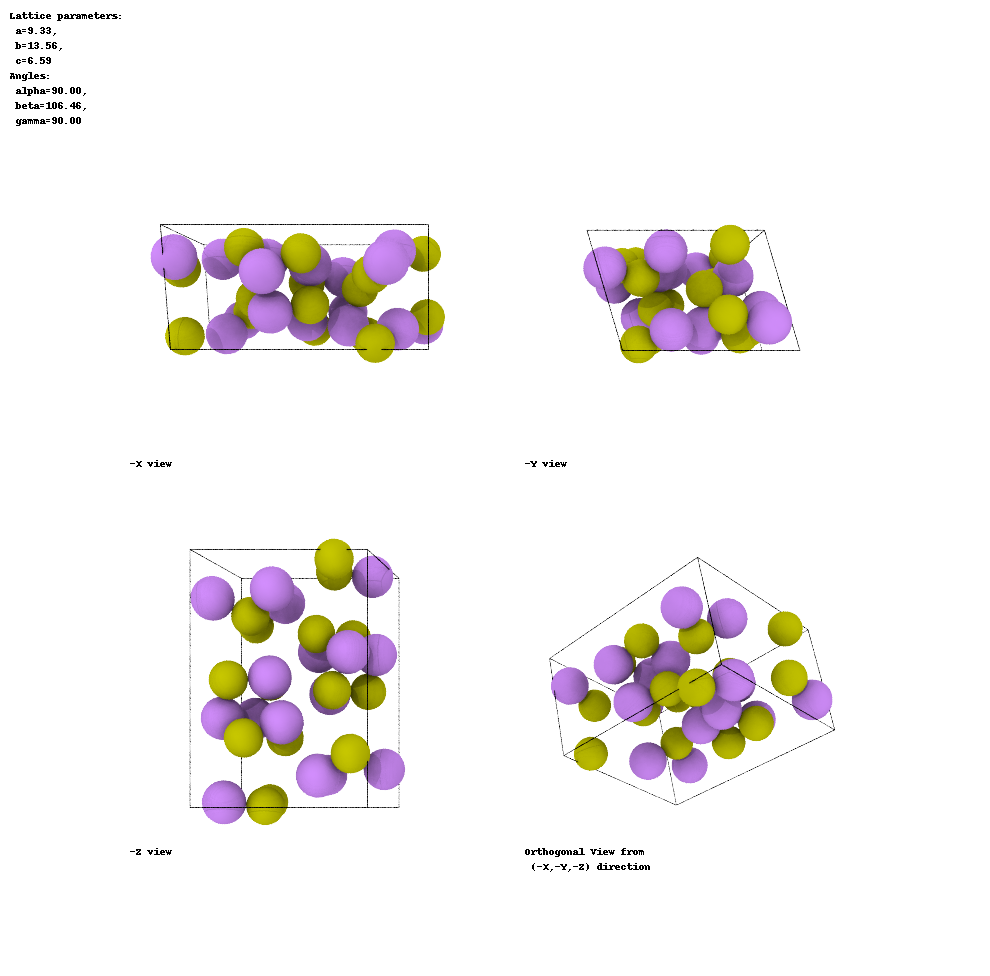}\par\smallskip
  {\footnotesize\raggedright
  \textbf{(c) Q:} What is the crystal system of the rendered unit
  cell?\par
  \textbf{MatVerse:} $a{\neq}b{\neq}c$ (8.33, 13.56, 6.59\,\AA),
  $\alpha{=}\gamma{=}90^{\circ}$, $\beta{=}106.46^{\circ}$ $\to$
  \textbf{monoclinic} (G).\par}
\end{minipage}\hfill
\begin{minipage}[t]{0.48\linewidth}
  \centering
  \includegraphics[width=0.62\linewidth]{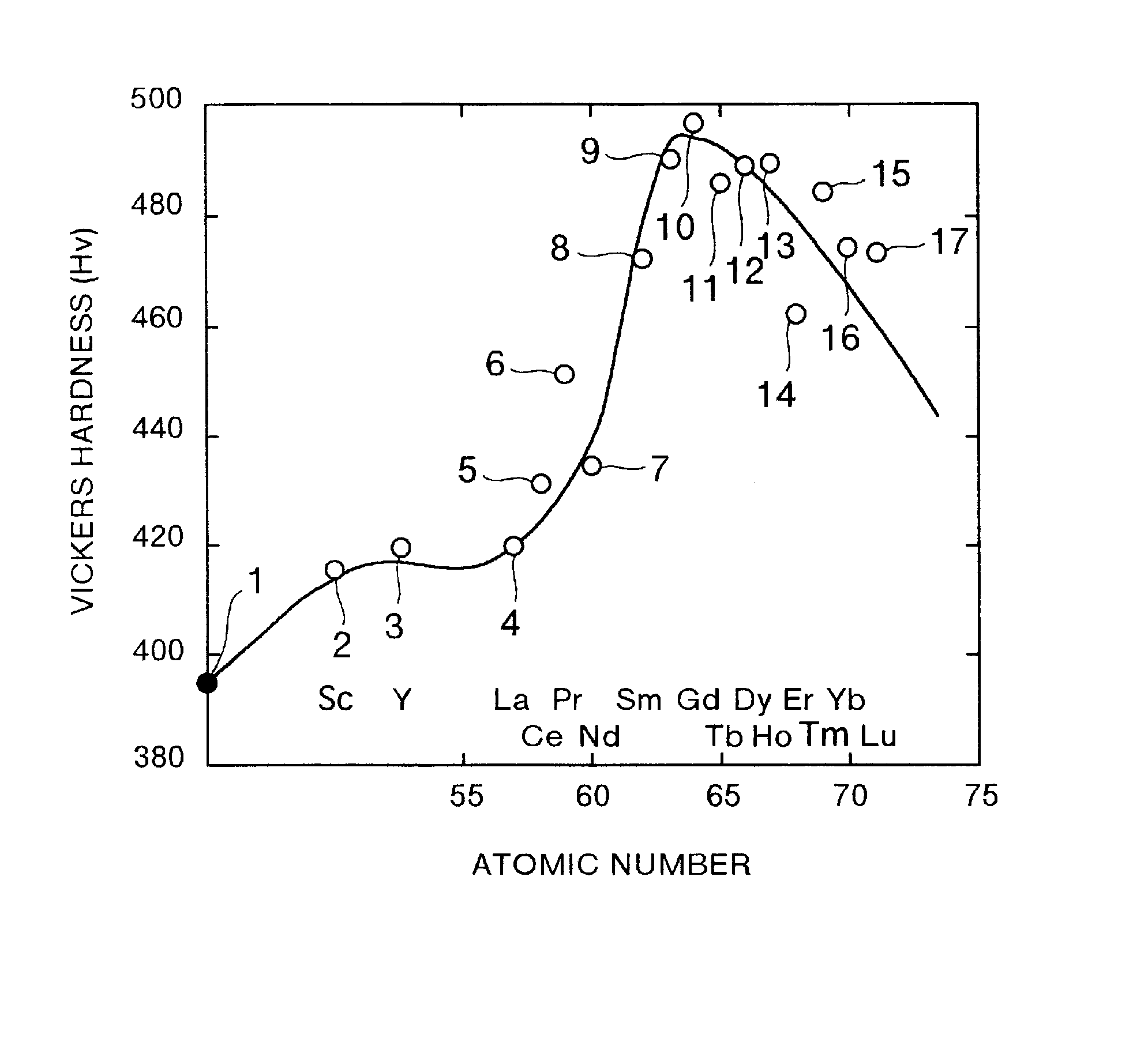}\par\smallskip
  {\footnotesize\raggedright
  \textbf{(d) Q:} What hardness (Hv) does sample~4 exhibit in the patent
  figure?\par
  \textbf{MatVerse:} The point labeled ``4'' lies exactly on the 420
  gridline of the Vickers-hardness axis $\to$ \textbf{420}.\par}
\end{minipage}
\par\medskip
\begin{minipage}[t]{0.48\linewidth}
  \centering
  \includegraphics[width=\linewidth]{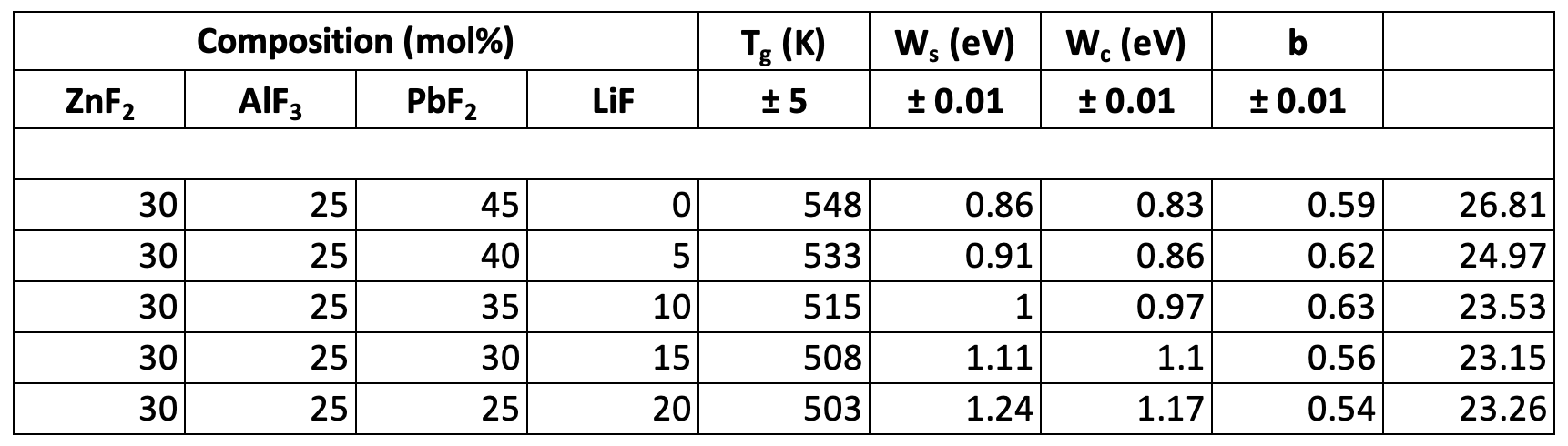}\par\smallskip
  {\footnotesize\raggedright
  \textbf{(e) Q:} How many unique compounds are present in the glass
  compositions?\par
  \textbf{MatVerse:} The ``Composition (mol\%)'' header lists four distinct
  compounds: ZnF$_2$, AlF$_3$, PbF$_2$, LiF $\to$ \textbf{4 compounds}.\par}
\end{minipage}\hfill
\begin{minipage}[t]{0.48\linewidth}
  \centering
  \includegraphics[width=0.88\linewidth]{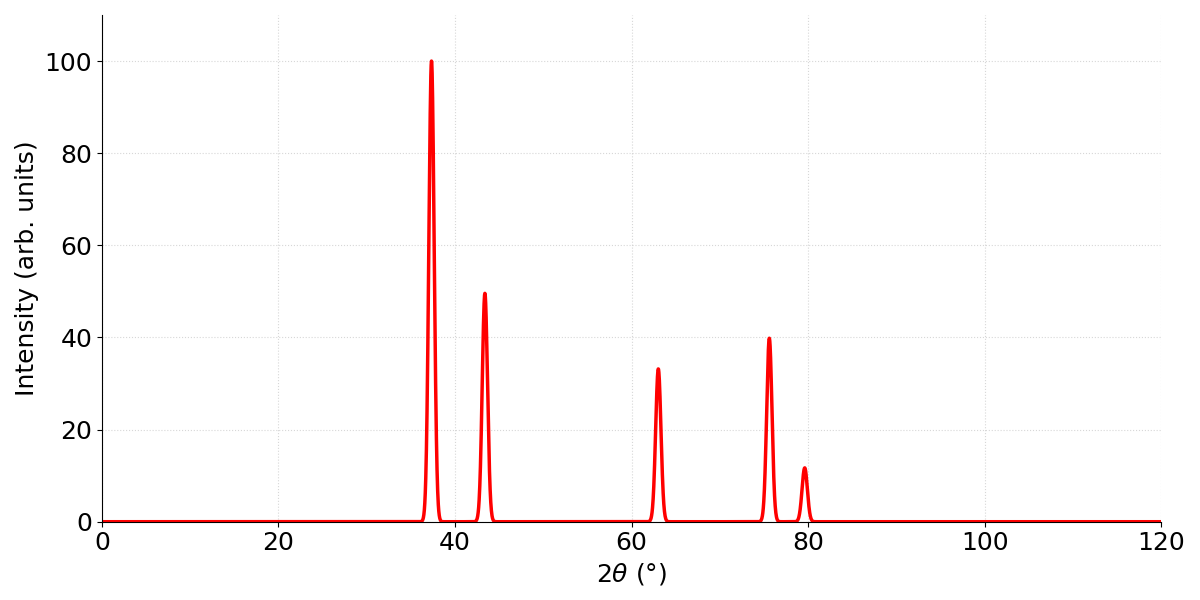}\par\smallskip
  {\footnotesize\raggedright
  \textbf{(f) Q:} Order the peak positions ($2\theta$) from highest to
  lowest intensity.\par
  \textbf{MatVerse:} $37.3^{\circ}$ (100), $43.4^{\circ}$ ($\sim$50),
  $75.6^{\circ}$ ($\sim$40) $\to$ \textbf{option C}
  ($37.3^{\circ}$, $43.4^{\circ}$, $75.6^{\circ}$).\par}
\end{minipage}
\caption{Qualitative question--answering examples of \method{} sampled from
the MaCBench evaluation (responses condensed; the model answers all six
questions correctly).
(a)~Hand-drawn molecule to SMILES.
(b)~Laboratory equipment recognition.
(c)~Crystal-system identification from a structure rendering.
(d)~Data extraction from a US-patent plot.
(e)~Materials property table understanding.
(f)~XRD relative-intensity analysis.
Full prompts, thinking trajectories, and complete responses are provided in
Appendix~\ref{app:macbench-cases}; condensed long-chain MatSciBench
reasoning cases are provided in
Appendix~\ref{app:matscibench-cases}.}
\label{fig:qualitative}
\end{figure}

\textbf{Qualitative examples.} Figure~\ref{fig:qualitative} presents
correctly answered multimodal questions sampled from the MaCBench evaluation
of \method{}, spanning hand-drawn molecule recognition, laboratory equipment
identification, crystal-system determination from structure renderings,
materials property table understanding, XRD peak analysis, and data
extraction from US-patent plots. In these cases, the model grounds
its answers in the provided visual evidence rather than generic priors, for
example by reading lattice parameters from a structure rendering, reading
peak positions and intensities from XRD patterns, and
locating labeled data points on patent-plot gridlines.
Interface-style cases on OmniMatBench are provided in
Appendix~\ref{app:qualitative-cases}, the full prompts, thinking
trajectories, and complete responses for
Figure~\ref{fig:qualitative} are provided in
Appendix~\ref{app:macbench-cases}, and condensed long-chain MatSciBench
reasoning traces---including a multi-step Frenkel defect calculation and a
materials-selection problem with embodied-energy analysis where the model
detects and corrects an initial estimation error---are provided in
Appendix~\ref{app:matscibench-cases}.

\section{Conclusion, Limitations, and Future Work}

\method{} is obtained by post-training Intern-S2-Preview on nearly 150K
materials-domain examples, targeting materials knowledge, quantitative
reasoning, and the interpretation of scientific visual data within a single
model. On MatSciBench, \RuleAcc{} improves from 58.45\% to 63.59\%. On
MaCBench, \OverAcc{} improves from 65.05\% to 67.22\%. On OmniMatBench,
\method{} attains 24.45\% \ESA{} on the CAL track, improving over the base
model, while achieving 27.59\% \MacroFOne{} on the QA track, slightly below
the base model. These results demonstrate clear gains on several
materials-science benchmarks, while showing that the benefits of
materials-specific post-training vary across evaluation tasks.

Several limitations remain. Open-ended question answering still trails the
strongest reported systems (27.59\% vs.\ 44.25\% \MacroFOne{} for
Claude~Opus~4.7 on OmniMatBench QA), formula derivation on MatSciBench stays
near 9\% \RuleAcc{}, and a few subfields remain close to zero \ESA{}. In the
present evaluation, \method{} also operates as a standalone generator without
external retrieval or tools. Future work will explore reinforcement learning
on verifiable calculation tasks, additional supervision for underrepresented
subfields, better preservation of general question-answering capability during
domain post-training, and agent integration with materials databases and
scientific tools.

\clearpage
\bibliographystyle{assets/plainnat}

\bibliography{reference_no_fig5ab}

\clearpage
\appendix
\begingroup
\setlength{\intextsep}{4pt}
\setlength{\parskip}{3pt}
\renewcommand{\arraystretch}{0.92}
\captionsetup{skip=3pt}
\section{Additional Details of Post-Training Data Construction}
\label{app:data}

\subsection{Reasoning Data Filtering Statistics}

For the synthetic reasoning data, each generated problem is independently solved ten times. After answer normalization, we retain a problem only if the majority answer receives at least six votes.

Table~\ref{tab:rollout_filtering} reports the intermediate filtering statistics.

\begin{table}[htbp]
\centering
\caption{Statistics of multi-rollout consistency filtering.}
\label{tab:rollout_filtering}
\small
\begin{tabular}{lr}
\toprule
\textbf{Statistic} & \textbf{Value} \\
\midrule
Synthetic questions before filtering & 5,938 \\
Samples passing majority filtering & 2,240 \\
Samples rejected by consistency filtering & 3,512 \\
Intermediate retention rate & 37.7\% \\
Samples with non-empty reasoning & 100\% \\
\bottomrule
\end{tabular}
\end{table}

The filtering rate varies considerably across materials subdomains. Problems dominated by relatively well-defined numerical calculations tend to exhibit higher agreement across rollouts, while tasks requiring longer reasoning chains or permitting broader answer spaces show lower self-consistency.

\subsection{Multimodal Retrieval Statistics}

For multimodal data construction, each MaCBench image is used as a query. We run text-to-image and image-to-image retrieval with depth $K=200$. Table~\ref{tab:retrieval_statistics} and Figure~\ref{fig:retrieval_scale} summarize the full candidate pool. These counts are not the training set. Question--answer synthesis uses only the both-pathway export described below.

\begin{table}[htbp]
\centering
\caption{Full dual-path retrieval pool. Counts are unique images, except the last overlap row, which counts query--image pairs.}
\label{tab:retrieval_statistics}
\small
\begin{tabular}{lr}
\toprule
\textbf{Statistic} & \textbf{Value} \\
\midrule
Unique images retrieved by text-to-image search & 304,504 \\
Unique images retrieved by image-to-image search & 263,443 \\
Images retrieved by both pathways & 86,749 \\
Same image retrieved for the same query by both pathways & 18,792 \\
Retrieval depth per pathway & 200 \\
\bottomrule
\end{tabular}
\end{table}

\vspace{-4pt}
\begin{figure}[H]
\centering
\includegraphics[width=0.60\linewidth]{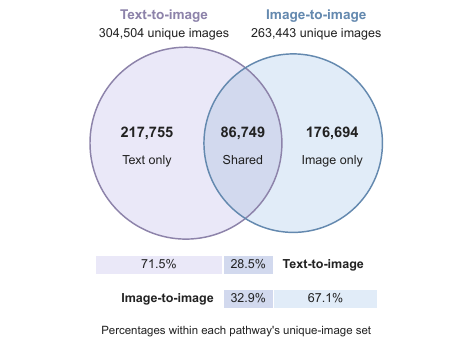}
\caption{Image-level Venn diagram of dual-path retrieval. Text-to-image and
image-to-image return 304,504 and 263,443 unique images, respectively; their
intersection contains 86,749 images ($217{,}755$ text-only and $176{,}694$
image-only). The bars below report the within-pathway shares of the
intersection. Same-query both-pathway matches ($18{,}792$) are listed in
Table~\ref{tab:retrieval_statistics} and are not drawn here.}
\label{fig:retrieval_scale}
\end{figure}

The two retrieval pathways exhibit different coverage and score
distributions. Text-to-image retrieval returns a larger set of candidate
images, with similarity scores concentrated below $0.85$, whereas
image-to-image retrieval returns fewer candidates but includes a substantial
fraction with scores above $0.85$, extending to $1.0$. The two pathways share
86,749 images at the image level. For 18,792 query--image pairs, the same
image is returned by both pathways for the same MaCBench query, indicating
that the majority of the image-level overlap corresponds to images retrieved
for different queries rather than exact same-query matches.

\begin{table}[htbp]
\centering
\caption{Statistics of the both-pathway subset used for multimodal
question--answer synthesis.}
\label{tab:retrieval_synthesis}
\small
\begin{tabular}{lr}
\toprule
\textbf{Statistic} & \textbf{Value} \\
\midrule
Both-pathway records linked to a MaCBench query & 52,922 \\
Unique images in that export & 52,902 \\
Records with score $\ge 0.60$ & 52,246 \\
Records removed by the score cutoff & 676 \\
Images with a successful generation & 51,757 \\
Question--answer pairs produced & 119,167 \\
Retained after format checks & 119,153 \\
\quad Multiple-choice & 111,598 \\
\quad Open-ended & 7,555 \\
\bottomrule
\end{tabular}
\end{table}

The subset used for question--answer synthesis is restricted to the
both-pathway records linked to MaCBench queries
(Table~\ref{tab:retrieval_synthesis}). Among the 52,922 records,
52,246 satisfy the retrieval-score threshold of $0.60$ and are retained for
synthesis. Figure~\ref{fig:retrieval_scores} shows the score distribution of
this subset; the 676 records below the threshold are excluded before
generation. The retained images yield 119,167 question--answer pairs, of
which 119,153 pass the subsequent format checks, including 111,598
multiple-choice and 7,555 open-ended examples.

\begin{figure}[H]
\centering
\includegraphics[width=0.8\linewidth]{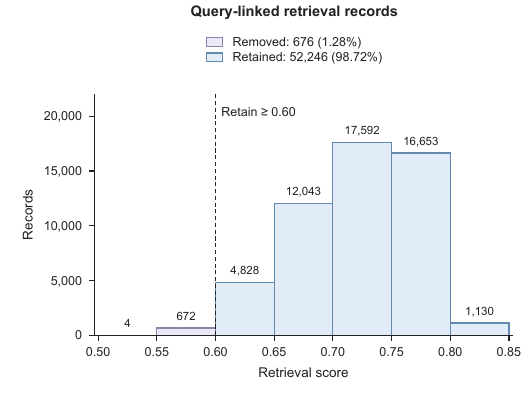}
\caption{Similarity-score distribution of the 52,922 both-pathway records
linked to MaCBench queries. The pale bars indicate the 676 records (1.28\%)
below the $0.60$ synthesis threshold, which are excluded before
question--answer generation. Image-only matches with scores above $0.85$ are
not included in this subset.}
\label{fig:retrieval_scores}
\end{figure}

Figures~\ref{fig:retrieval_scale} and~\ref{fig:retrieval_scores} summarize
the retrieval statistics and the actual both-pathway subset used for
synthesis. They are reconstructed from the retrieval outputs rather than
from individual retrieval-interface screenshots, which are omitted for
clarity.

\endgroup

\section{Hyperparameter Settings}
\label{app:hyperparameters}

Table~\ref{tab:hyperparameters} summarizes the hyperparameter settings of the
4B pilot experiments described in the main text. For the final 35B-A3B model,
the training configuration follows the large-scale training paradigm of
Intern-S2-Preview~\cite{bai2026intern}, together with the validated
materials-domain data recipe.

\begin{table}[htbp]
\centering
\caption{Hyperparameter settings for the 4B pilot training of \method{}.}
\label{tab:hyperparameters}
\small
\begin{tabular}{ll}
\toprule
\textbf{Hyperparameter} & \textbf{Value} \\
\midrule
Post-training objective & Supervised fine-tuning (SFT) \\
Trainable components & Language model (full parameter) \\
Frozen components & Visual encoder, multimodal alignment modules \\
Optimizer & Distributed AdamW-style optimizer \\
Peak learning rate & $1\times10^{-5}$ \\
Minimum learning rate & $1\times10^{-6}$ \\
Learning-rate warmup & 5\% \\
Gradient clipping & Max norm 1.0 \\
Global batch size & 32 \\
Maximum sequence length & 40,960 tokens \\
Sequence packing & Enabled \\
Training framework & Megatron-based ms-swift \\
Hardware & 256 Ascend NPUs \\
\bottomrule
\end{tabular}
\end{table}

\section{Prompts for Post-Training Data Construction}
\label{app:prompts}

For reproducibility, we summarize the major prompts used in our post-training data construction pipeline.
We group them according to the corresponding stage of data construction:
(1) scientific problem synthesis,
(2) solution rollout,
(3) reasoning distillation,
and (4) multimodal reasoning distillation.
Some prompts use slightly different answer-format constraints for different data sources; we report these differences explicitly below.

\subsection{Stage I: Scientific Problem Synthesis}
\label{app:prompt_synthesis}

We first expand the reasoning-oriented training corpus by generating new materials-science problems from seed examples.
The synthesis prompt preserves the original topic, problem type, reasoning pattern, and expected answer structure, while modifying problem-specific details such as numerical values, materials, units, physical conditions, or symbolic variables when scientifically appropriate.

We use the following general synthesis instruction.

\Needspace{60pt}
\begin{Prompt}
Help the user create new materials science problems similar to a given one.
Keep the topic, reasoning pattern, and answer structure similar.

Ensure every new problem is reasonable, solvable, and consistent with
established principles of materials science.
\end{Prompt}

Depending on the problem type, we apply one of the following task-specific instructions.

\paragraph{Numerical problems.}

\Needspace{36pt}
\begin{Prompt}
Primarily change numerical values, materials, units, and conditions
while keeping the same problem type, reasoning path, and answer-slot structure.
\end{Prompt}

\Needspace{7\baselineskip}
\paragraph{Formula-oriented problems.}

\Needspace{36pt}
\begin{Prompt}
You may change formula symbols, material context, crystallographic
directions or planes, or constants when doing so remains physically meaningful.
\end{Prompt}

The full synthesis template is:

\Needspace{252pt}
\begin{Prompt}
{template_instruction}

Ensure every new problem is reasonable, solvable, and consistent with
established principles of materials science.

Here are examples from the same benchmark distribution:

{few_shot_blocks}

Here is the problem from the user:
{seed_question}

Generation requirements:
- Generate exactly {num_variants} new problem statements.
- Only generate problem statements; do not generate answers,
  derivations, hints, labels, or explanations.
- Preserve the original task type and expected answer-slot structure.
- Expected answer format structure for reference only:
  {answer_format}
- Do not introduce image-dependent information unless the seed problem
  already requires an image.
- Do not copy the seed problem verbatim.
- Return ONLY valid JSON in the following format:

{
  "variants": [
    {"question": "..."}
  ]
}
\end{Prompt}

The corresponding system prompt is:

\Needspace{36pt}
\begin{Prompt}
You are a careful materials science dataset generator.
Return only valid JSON.
\end{Prompt}

\subsection{Stage II: Solution Rollout}
\label{app:prompt_rollout}

After problem synthesis, we generate candidate solutions and reasoning trajectories with a strong teacher model.
Across different data sources, the rollout prompt follows the same general structure:
the model is instructed to reason step by step and return a final answer in a machine-parsable format.

The common system instruction is:

\Needspace{92pt}
\begin{Prompt}
You are a materials science expert.
Your task is to provide a clear and accurate answer.

Reason step by step and provide both:
1. the complete reasoning process;
2. the final answer in the required output format.

Do not include additional commentary outside the required reasoning
and final-answer fields.
\end{Prompt}

\Needspace{125pt}
For examples that use structured answer slots, we adopt the following output constraint:

\Needspace{92pt}
\begin{Prompt}
Put all reasoning inside <think>...</think>.

Put ONLY the final answer inside <answer>...</answer>.

If a `final_answer_format` is provided, follow its grouping,
ordering, and number of answer slots exactly.

Replace each empty slot with exactly one final answer string containing
only the final number, formula, or expression.
\end{Prompt}

For quantitative problems evaluated with a single boxed answer, the output constraint is simplified to:

\Needspace{52pt}
\begin{Prompt}
Please reason step by step and put the final answer inside a single
box using \boxed{...}.

Include only the final answer inside the box, without the unit.
\end{Prompt}

When a few-shot example is used, the user-side template follows the common form:

\Needspace{124pt}
\begin{Prompt}
Here are some examples of questions and solutions followed by a new
question that you need to solve.

Question:
{raw_question}

My solution:
{raw_solution}

Question:
{question}

My solution:
\end{Prompt}

For quantitative tasks, an additional instruction is appended:

\Needspace{28pt}
\begin{Prompt}
Make sure to put the answer, and only the answer, inside \boxed{}.
\end{Prompt}

For problems whose expected unit is known, the unit is appended to the
question as auxiliary information while the model is instructed not to
include the unit inside the final boxed answer.

\Needspace{220pt}
\subsection{Stage III: Reasoning Distillation}
\label{app:prompt_distillation}

For samples with known correct answers, we additionally construct reasoning supervision by asking the teacher model to reconstruct a complete solution trajectory.
The correct answer and, when available, reference key points are provided only as hidden guidance.
The generated response is required to resemble a solution derived from scratch.

The teacher system prompt is:

\Needspace{180pt}
\begin{Prompt}
You are a materials science expert helping create high-quality
reasoning training data.

You will receive a question, optional reference key points,
and the correct final answer.

Your job is to write a complete step-by-step reasoning chain as if
you are solving the problem from scratch, without any prior knowledge
of the answer.

Strict requirements:
1. Do NOT mention that the answer or key points were given to you.
2. Do NOT use phrases such as "given the answer", "provided answer",
   "we know the result is", or "according to the key points".
3. Use the key points only as hidden guidance for the logical flow.
4. Show the derivation explicitly when appropriate, including knowns,
   unknowns, formulas, substitutions, and calculations.
5. Put all reasoning inside <think>...</think>.
6. Put ONLY the final answer inside <answer>...</answer>.
7. Do not add text outside the reasoning and answer blocks.
\end{Prompt}

\Needspace{15\baselineskip}
The corresponding user template is:

\Needspace{108pt}
\begin{Prompt}
Question:
{question}

Correct final answer:
{final_answer}

Reference key points (optional; hidden guidance only):
- {key_point_1}
- ...

Now write the full reasoning chain and final answer in the required format.
\end{Prompt}

For data sources that do not provide reference key points, the same template is used with only the question and the correct final answer.

\subsection{Stage IV: Multimodal Reasoning Distillation}
\label{app:prompt_multimodal_cot}

For selected multimodal question-answer examples with known final answers, we reconstruct reasoning trajectories from the question, image context, and final answer.
The prompt is language-adaptive: an English or Chinese version is selected according to the language of the input question.

\paragraph{English template.}

\Needspace{156pt}
\begin{Prompt}
You need to reconstruct the full reasoning process from the question
to the final answer.

Requirements:
1. Follow a natural reasoning style.
2. Reason step by step without skipping key steps.
3. Keep the reasoning coherent with the question and available visual context.
4. Output the reasoning chain first, then give the final answer.

Question:
{question}

Final answer:
{answer}

Please output:
[Reasoning Chain] + [Final Answer]
\end{Prompt}

\paragraph{Chinese template.}

\Needspace{148pt}
\begin{Prompt}
你需要还原从问题推理到最终答案的完整思考过程。

要求：
1. 按照自然的推理方式组织思考过程；
2. 一步步拆解，不跳过关键步骤；
3. 推理过程应与问题及可用的图像信息保持一致；
4. 先输出思维链，再给出最终答案。

问题：
{question}

最终答案：
{answer}

请输出：
【思维链】+【最终答案】
\end{Prompt}

\subsection{Stage V: Consistency Filtering and Data Selection}
\label{app:prompt_filtering}

The consistency filtering stage does not introduce an additional prompt.
Instead, it operates on multiple independently sampled rollout results.

For each synthesized problem, we generate ten candidate solutions.
The final answers are normalized and grouped by equivalence.
A problem is retained only when the majority answer receives at least
six votes out of ten.
Among the rollouts associated with the majority answer, we preferentially
select one with a complete and non-empty reasoning trajectory.

This procedure is applied after solution generation and before the
examples are converted into the final supervised fine-tuning format.

\section{Qualitative Case Studies}
\label{app:qualitative-cases}

\subsection{Molecular String}
\Needspace{10\baselineskip}
\CaseTitle{BBBP: Propanolol BBB penetration}

\begin{OmniUserBox}
\begin{Verbatim}[breaklines=true,breakanywhere=true,fontsize=\scriptsize]
[MATERIAL_TYPE] molecule
[SMILES] [Cl].CC(C)NCC(O)COc1cccc2ccccc12
\end{Verbatim}
\vspace{0.2em}
该分子是否能穿透血脑屏障（BBB）？请回答能或不能，并简要说明。
\end{OmniUserBox}

\begin{OmniModelBox}
\texttt{\textless think\textgreater}\\
该分子（普萘洛尔）具有脂溶性，分子量较小（约260），且含有一个碱性胺基（在生理pH下部分带电荷），其理化性质符合穿透血脑屏障的特征，因此能够进入中枢神经系统。\\
\texttt{\textless /think\textgreater}\\[0.35em]
\\[-0.2em]
最终答案：能
\end{OmniModelBox}

\Needspace{0.45\textheight}
\subsection{2D Molecular Structure}
\Needspace{10\baselineskip}
\CaseTitle{BBBP 2D: SKF-93619 BBB penetration}

\begin{OmniUserBox}
\begin{center}
\includegraphics[width=0.58\linewidth]{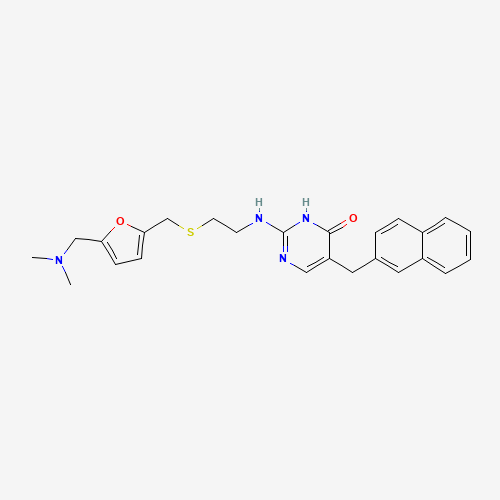}
\end{center}
该分子是否能穿透血脑屏障（BBB）？请回答能或不能，并简要说明。
\end{OmniUserBox}

\begin{OmniModelBox}
\texttt{\textless think\textgreater}\\
该分子具有中等大小的分子量（约425 Da），含有多个氢键供体和受体（如氨基、羰基、呋喃环上的氧原子等），导致其极性较大。此外，分子中含有一个叔胺基团，在生理pH下会带正电荷。这些高极性和带电的特性使其难以通过被动扩散穿过血脑屏障。\\
\texttt{\textless /think\textgreater}\\[0.35em]
\\[-0.2em]
因此，答案是 不能。
\end{OmniModelBox}

\Needspace{0.75\textheight}
\subsection{Multi-view 3D Structure}
\Needspace{10\baselineskip}
\CaseTitle{COD 9008565: tetrahedral geometry}

\begin{OmniUserBox}
\begin{Verbatim}[breaklines=true,breakanywhere=true,fontsize=\scriptsize]
5
COD 9008565 local coordination
Si 0.0000 0.0000 0.0000
Si -1.3577 -1.3577 1.3577
Si -1.3577 1.3577 -1.3577
Si 1.3577 -1.3577 -1.3577
Si 1.3577 1.3577 1.3577
\end{Verbatim}
\vspace{0.2em}
\begin{center}
\begin{minipage}[t]{0.47\linewidth}
\centering
\includegraphics[width=0.55\linewidth]{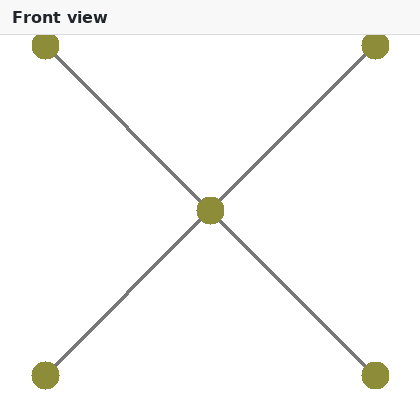}\\[-0.2em]
{\scriptsize 视角：正视图（front）}
\end{minipage}\hfill\begin{minipage}[t]{0.47\linewidth}
\centering
\includegraphics[width=0.55\linewidth]{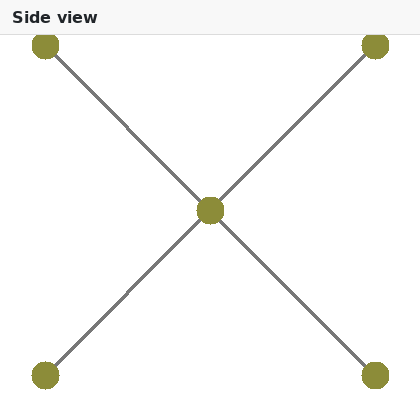}\\[-0.2em]
{\scriptsize 视角：侧视图（side）}
\end{minipage}\\[0.8em]
\begin{minipage}[t]{0.47\linewidth}
\centering
\includegraphics[width=0.55\linewidth]{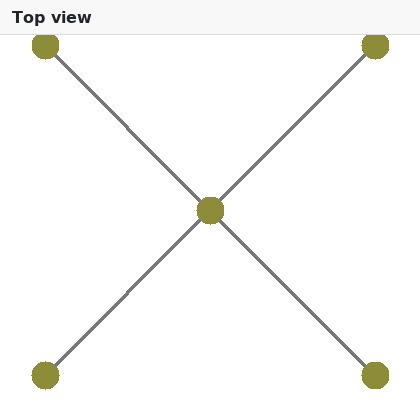}\\[-0.2em]
{\scriptsize 视角：俯视图（top）}
\end{minipage}\hfill\begin{minipage}[t]{0.47\linewidth}
\centering
\includegraphics[width=0.55\linewidth]{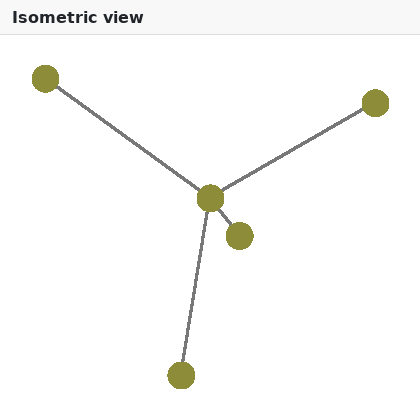}\\[-0.2em]
{\scriptsize 视角：等轴测视图（iso）}
\end{minipage}\\[0.8em]
\end{center}
根据多视角结构图和 XYZ 坐标，判断中心配位原子的配位几何类型。
\end{OmniUserBox}
\begin{OmniModelBox}
\texttt{\textless think\textgreater}\\
中心原子位于原点 (0,0,0)，周围有4个配位原子。观察等轴测视图，这4个配位原子呈四面体分布（类似甲烷或金刚石结构中的局部配位）。XYZ坐标也符合四面体构型（例如，四个顶点位于交替的卦限中）。\\
\texttt{\textless /think\textgreater}\\[0.35em]
\\[-0.2em]
因此，答案是 四面体 (Tetrahedral)。\\
\\[-0.2em]
\\[-0.2em]
四面体 (Tetrahedral)
\end{OmniModelBox}

\Needspace{0.4\textheight}
\subsection{Crystal Structure}
\Needspace{10\baselineskip}
\CaseTitle{MgO periclase (COD)}

\begin{OmniUserBox}
\begin{Verbatim}[breaklines=true,breakanywhere=true,fontsize=\scriptsize]
data_1000053
_chemical_formula_structural     'Mg O'
_chemical_formula_sum            'Mg O'
_chemical_name_mineral           Periclase
_chemical_name_systematic        'Magnesium oxide'
_space_group_IT_number           225
_symmetry_cell_setting           cubic
_symmetry_Int_Tables_number      225
_symmetry_space_group_name_Hall  '-F 4 2 3'
_symmetry_space_group_name_H-M   'F m -3 m'
_cell_angle_alpha                90
_cell_angle_beta                 90
_cell_angle_gamma                90
_cell_formula_units_Z            4
_cell_length_a                   4.217(1)
_cell_length_b                   4.217(1)
_cell_length_c                   4.217(1)
_cell_volume                     75.0
_cod_database_code               1000053
loop_
_atom_site_label
_atom_site_type_symbol
_atom_site_symmetry_multiplicity
_atom_site_Wyckoff_symbol
_atom_site_fract_x
_atom_site_fract_y
_atom_site_fract_z
_atom_site_occupancy
Mg1 Mg2+ 4 a 0. 0. 0. 1.
O1 O2- 4 b 0.5 0.5 0.5 1.
...
(symmetry-equivalent positions omitted in compact version)
\end{Verbatim}
\vspace{0.2em}
\begin{center}
\includegraphics[width=0.34\linewidth]{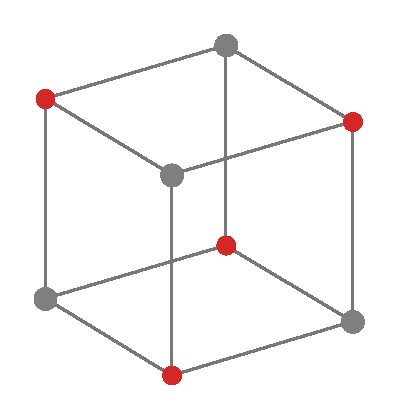}
\\[-0.1em]
{\scriptsize 视角：晶胞示意渲染（iso）}
\end{center}
根据 CIF 文本和晶胞渲染图，判断该晶体结构类型、空间群或典型配位环境。
\end{OmniUserBox}

\begin{OmniModelBox}
\texttt{\textless think\textgreater}\\
CIF 文本明确指出该结构为氧化镁（Periclase），空间群为 F m -3 m（225 号），属于典型的面心立方（NaCl 型）结构。渲染图中展示的立方体晶胞中，红灰两色原子分别占据顶点和面心位置（或棱心和体心），形成相互穿插的两个面心立方格子，每个原子的配位数为 6（八面体配位）。\\
\texttt{\textless /think\textgreater}\\[0.35em]
\\[-0.2em]
最终答案：NaCl型结构，空间群F m -3 m，配位数为6。
\end{OmniModelBox}

\subsection{Characterization Spectrum}
\Needspace{10\baselineskip}
\CaseTitle{RRUFF XRD: Calcite}

\begin{OmniUserBox}
\begin{Verbatim}[breaklines=true,breakanywhere=true,fontsize=\scriptsize]
2theta,intensity
23.0900,8.36
29.4400,100.00
36.0200,14.36
39.4600,18.81
43.2200,14.64
47.1800,6.50
47.5800,19.73
48.5800,20.46
57.4800,9.78
60.7600,5.68
64.7600,6.70
65.7200,3.76
\end{Verbatim}
\vspace{0.2em}
\begin{center}
\includegraphics[width=0.66\linewidth]{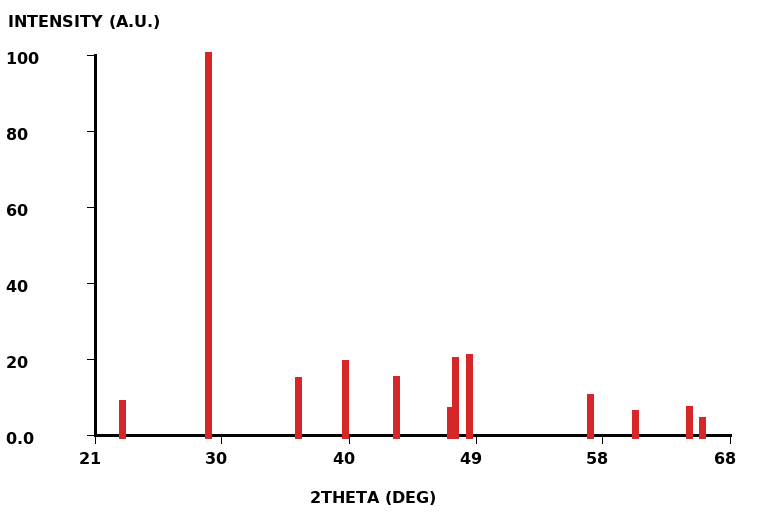}
\end{center}
根据 XRD 谱图和峰位表，判断可能的矿物/物相，并简要说明依据。
\end{OmniUserBox}

\begin{OmniModelBox}
\texttt{\textless think\textgreater}\\
谱图的最强特征峰位于 $2\theta \approx 29.44^\circ$（相对强度 100\%），这是方解石（Calcite）最典型的衍射峰。此外，在 $2\theta \approx 39.5^\circ$ 和 $47.6^\circ$--$48.6^\circ$ 附近出现的次强峰簇也与方解石的标准衍射数据高度吻合。\\
\texttt{\textless /think\textgreater}\\[0.35em]
\\[-0.2em]
因此，答案是 方解石 (Calcite)。\\
\\[-0.2em]
\\[-0.2em]
方解石 (Calcite)
\end{OmniModelBox}

\Needspace{0.45\textheight}
\subsection{Microscopy Image}
\Needspace{10\baselineskip}
\CaseTitle{SEM of spherical AgCl nanoparticles}

\begin{OmniUserBox}
\begin{center}
\includegraphics[width=0.86\linewidth]{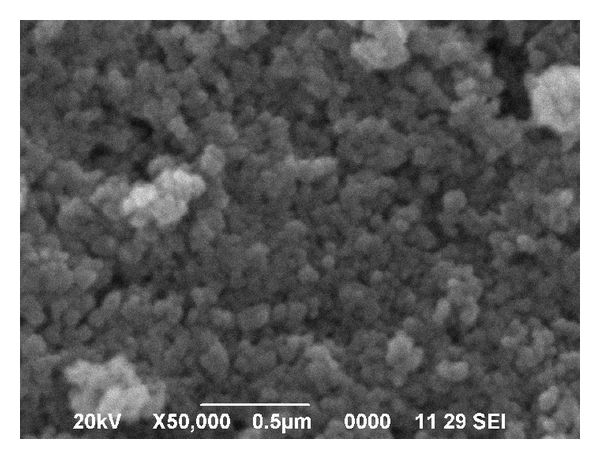}
\end{center}
这是一张 SEM 图。请描述微观形貌，并判断颗粒形状、粒径是否均匀、是否有团聚。
\end{OmniUserBox}

\begin{OmniModelBox}
\texttt{\textless think\textgreater}\\
图像显示大量不规则形状的纳米颗粒紧密堆积。颗粒尺寸分布相对均匀，但存在明显的团聚现象，形成了较大的聚集体，且颗粒间存在不规则的孔隙。\\
\texttt{\textless /think\textgreater}\\[0.35em]
\\[-0.2em]
最终答案：颗粒形状不规则，粒径相对均匀，存在团聚现象。
\end{OmniModelBox}

\Needspace{0.45\textheight}
\subsection{Process / Property Curve}
\Needspace{10\baselineskip}
\CaseTitle{MINICOR TGA: Beech}

\begin{OmniUserBox}
\begin{Verbatim}[breaklines=true,breakanywhere=true,fontsize=\scriptsize]
Temperature_C,Mass_percent
29.63,100.852
100.67,102.568
148.89,102.252
199.35,102.076
225.06,101.361
249.57,99.811
273.89,95.028
300.74,82.951
325.27,70.458
349.95,53.469
374.67,40.912
399.46,38.557
449.17,35.250
498.92,33.413
548.80,32.110
600.29,30.209
...
(full CSV contains 289 lines; omitted here for compact appendix layout)
\end{Verbatim}
\vspace{0.2em}
\begin{center}
\includegraphics[width=0.76\linewidth]{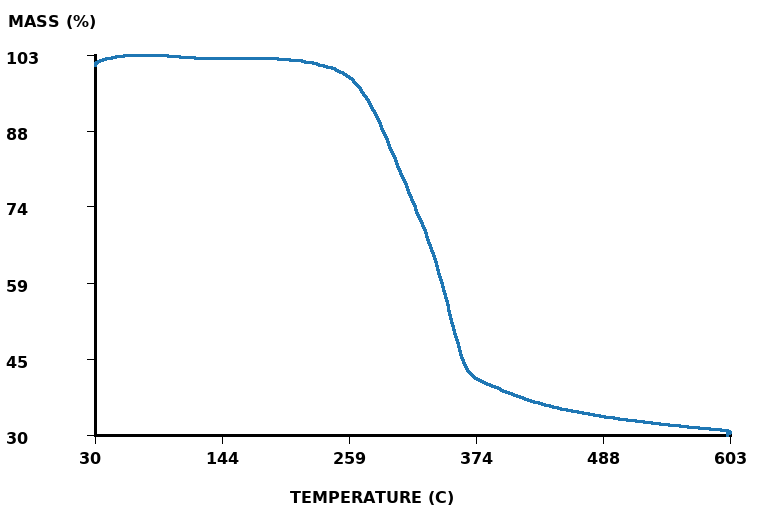}
\end{center}
这是 EU MINICOR 项目的真实 TGA 曲线。请根据曲线形态描述主要失重/热事件区间，并判断它更可能对应哪一类生物质样品（山毛榉 Beech / 松木 Pine / 杨木 Poplar 等）。
\end{OmniUserBox}

\begin{OmniModelBox}
\texttt{\textless think\textgreater}\\
该曲线是典型的热重分析（TGA）曲线。样品在约 250°C 之前保持稳定（甚至因氧化略有增重），随后经历剧烈的质量损失，主要分解区间在 250°C 至 375°C 之间，之后缓慢分解至约 30\% 的残余质量。这种在 300°C 左右发生剧烈分解的特征是半纤维素和纤维素热解的典型表现，符合山毛榉（Beech）等阔叶木的热解行为。\\
\texttt{\textless /think\textgreater}\\[0.35em]
\\[-0.2em]
因此，答案是 Beech。
\end{OmniModelBox}

\subsection{MaCBench: Multimodal Question Answering}
\label{app:macbench-cases}

The following cases are drawn from the MaCBench evaluation of \method{}, where
the model answers real English multimodal questions correctly. Responses are
condensed; the reasoning is quoted verbatim from the model outputs.

\Needspace{10\baselineskip}
\CaseTitle{Hand-drawn structure to SMILES (MaCBench)}

\begin{OmniUserBox}
\begin{center}
\includegraphics[width=0.48\linewidth]{macbench_hand_drawn_ps_ring.png}
\end{center}
What is a valid SMILES representation for the molecule with the hand-drawn
structure in the image?\\
{\scriptsize
A.~\texttt{CSP1(=S)SP(=S)(S1)SC}\quad
B.~\texttt{C/C=C\textbackslash C(=N/C(=O)NO)\textbackslash C(C)(C)C}\quad
C.~\texttt{CC(C)(C)C1(SSSS1)C(C)(C)C}\quad
D.~\texttt{CC(C1=PC=PC=P1)O}}
\end{OmniUserBox}

\begin{OmniModelBox}
The image shows a cyclic molecule with two phosphorus (P) atoms and three
sulfur (S) atoms forming a ring. Each phosphorus atom is double-bonded to a
sulfur atom (=S) and single-bonded to a methylthio group
(\texttt{-S-CH$_3$}). \texttt{CSP1(=S)SP(=S)(S1)SC} correctly maps to this
connectivity.\\[0.35em]
Final answer: \textbf{A}
\end{OmniModelBox}

\Needspace{0.65\textheight}
\CaseTitle{Laboratory equipment recognition (MaCBench)}

\begin{OmniUserBox}
\begin{center}
\includegraphics[width=0.30\linewidth]{macbench_round_flask.png}
\end{center}
Which equipment from a chemistry lab do you see in this image?\\
{\scriptsize
A.~Chromatography Column\quad B.~Round Flask\quad C.~Beaker\quad
D.~Separatory Funnel}
\end{OmniUserBox}

\begin{OmniModelBox}
The image shows a round-bottom flask, which is identifiable by its spherical
body and narrow neck. It is commonly used for heating or mixing chemicals,
unlike beakers (cylindrical), separatory funnels (teardrop-shaped with a
stopcock), or chromatography columns (tall tubes).\\[0.35em]
Final answer: \textbf{B} (Round Flask)
\end{OmniModelBox}

\Needspace{10\baselineskip}
\CaseTitle{Materials property table understanding (MaCBench)}

\begin{OmniUserBox}
\begin{center}
\includegraphics[width=0.72\linewidth]{macbench_glass_table.png}
\end{center}
The table shows compositions, activation energies for the conductivity and
the relaxation time, stretching exponent, and high-frequency dielectric
constant for the ZnF$_2$--AlF$_3$--PbF$_2$--LiF glasses. How many unique
chemical compounds are present in the glass compositions described here?
\end{OmniUserBox}

\begin{OmniModelBox}
The table header under ``Composition (mol\%)'' lists four distinct chemical
compounds: ZnF$_2$, AlF$_3$, PbF$_2$, and LiF.\\[0.35em]
Final answer: \textbf{4}
\end{OmniModelBox}

\Needspace{10\baselineskip}
\CaseTitle{XRD peak-position reading (MaCBench)}

\begin{OmniUserBox}
\begin{center}
\includegraphics[width=0.72\linewidth]{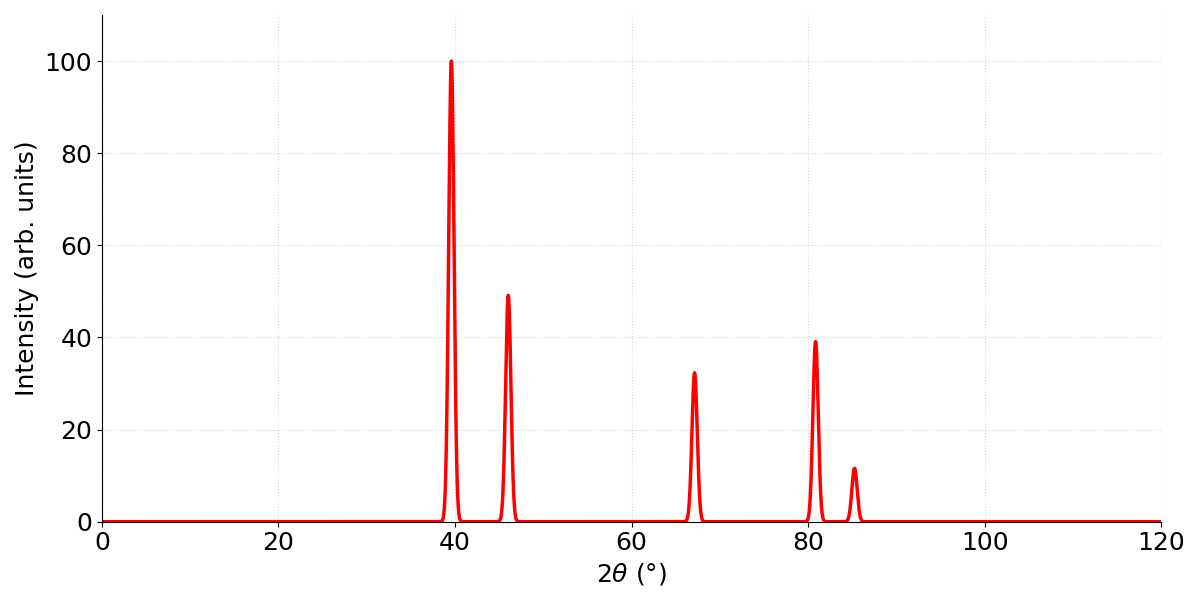}
\end{center}
What is the position ($2\theta$ value) of the most intense peak in the XRD
pattern shown in the image?\\
{\scriptsize A.~70.0\quad B.~68.3\quad C.~69.7\quad D.~39.6}
\end{OmniUserBox}

\begin{OmniModelBox}
The most intense peak in the XRD pattern is the one reaching an intensity of
100 arb.\ units on the y-axis. Looking at the x-axis, this peak is located
just before the $40^{\circ}$ mark, at approximately $39.6^{\circ}$.\\[0.35em]
Final answer: \textbf{D}
\end{OmniModelBox}

\Needspace{10\baselineskip}
\CaseTitle{XRD relative-intensity ordering (MaCBench)}

\begin{OmniUserBox}
\begin{center}
\includegraphics[width=0.72\linewidth]{macbench_xrd_intensity.png}
\end{center}
What is the correct order of peak positions ($2\theta$) from highest to
lowest intensity in the XRD pattern shown in the image?\\
{\scriptsize
A.~43.4, 75.6, 37.3\quad
B.~37.3, 75.6, 43.4\quad
C.~37.3, 43.4, 75.6\quad
D.~43.4, 37.3, 75.6}
\end{OmniUserBox}

\begin{OmniModelBox}
The XRD pattern shows the highest intensity peak at
$2\theta\approx37.3^{\circ}$ (100 arb.\ units), the second highest at
$2\theta\approx43.4^{\circ}$ ($\sim$50 arb.\ units), and the third highest
at $2\theta\approx75.6^{\circ}$ ($\sim$40 arb.\ units). Therefore, the
order from highest to lowest intensity is 37.3, 43.4, 75.6.\\[0.35em]
Final answer: \textbf{C}
\end{OmniModelBox}

\Needspace{10\baselineskip}
\CaseTitle{Data extraction from a US-patent plot (MaCBench)}

\begin{OmniUserBox}
\begin{center}
\includegraphics[width=0.56\linewidth]{macbench_patent_hv.png}
\end{center}
Fig.~2 in the image is a graph that shows the relation between the atomic
number of the rare-earth elements contained in the glass and the micro
Vickers hardness of the glass. What is the hardness in Hv exhibited by
sample~4?
\end{OmniUserBox}

\begin{OmniModelBox}
The data point labeled ``4'' is located exactly on the horizontal grid line
corresponding to 420 on the Vickers-hardness (Hv) y-axis.\\[0.35em]
Final answer: \textbf{420}
\end{OmniModelBox}

\Needspace{10\baselineskip}
\CaseTitle{Crystal-system identification from a structure rendering
(MaCBench)}

\begin{OmniUserBox}
\begin{center}
\includegraphics[width=0.5\linewidth]{macbench_crystal_monoclinic.png}
\end{center}
This is a rendering of a crystal structure. What is the crystal system of
the unit cell?\\
{\scriptsize
A.~rhombohedral\quad B.~triclinic\quad C.~cubic\quad D.~orthorhombic\quad
E.~hexagonal\quad F.~tetragonal\quad G.~monoclinic}
\end{OmniUserBox}

\begin{OmniModelBox}
The image explicitly provides the lattice parameters and angles in the
top-left corner: $a=8.33$~\AA, $b=13.56$~\AA, $c=6.59$~\AA{} and
$\alpha=90.00^{\circ}$, $\beta=106.46^{\circ}$, $\gamma=90.00^{\circ}$.
The condition $a\neq b\neq c$, $\alpha=\gamma=90^{\circ}$, and
$\beta\neq90^{\circ}$ uniquely defines the monoclinic crystal
system.\\[0.35em]
Final answer: \textbf{G} (monoclinic)
\end{OmniModelBox}

\subsection{MatSciBench: Long-Chain Quantitative Reasoning}
\label{app:matscibench-cases}

The following cases illustrate the multi-step quantitative reasoning of
\method{} on MatSciBench numerical problems (roughly 4--5K generated tokens
each). Responses are condensed from the model outputs.

\Needspace{10\baselineskip}
\CaseTitle{Frenkel defect concentration in AgBr (MatSciBench)}

\begin{OmniUserBox}
Estimate the number of Frenkel defects in AgBr (NaCl structure) at
$500^{\circ}$C. The enthalpy of formation of the defect is 110~kJ/mol, and
the entropy of formation is $6.6R$. The density and molecular weight are
6.5~g/cm$^3$ and 187.8~g/mol, respectively. Assume that Frenkel disorder
occurs on the cation sublattice and that the silver ions occupy tetrahedral
interstitial sites. The unit of the answer is defects/cm$^3$.
\end{OmniUserBox}

\begin{OmniModelBox}
\texttt{\textless think\textgreater}\\
The number of Frenkel defects is
$n=\sqrt{NN_i}\,\exp[-\Delta G_f/(2RT)]$ with
$\Delta G_f=\Delta H_f-T\Delta S_f$.\\
(1)~Cation site density:
$N=\rho N_A/M=(6.5\times6.022\times10^{23})/187.8
=2.084\times10^{22}$~cm$^{-3}$.\\
(2)~Tetrahedral interstitial sites: $N_i=2N=4.169\times10^{22}$~cm$^{-3}$.\\
(3)~Free energy of formation:
$\Delta G_f=110{,}000-773.15\times6.6\times8.314
=6.576\times10^{4}$~J/mol.\\
(4)~Boltzmann factor:
$\exp[-\Delta G_f/(2RT)]=e^{-5.257}=5.21\times10^{-3}$.\\
(5)~Defect density:
$n=\sqrt{NN_i}\,e^{-\Delta G_f/(2RT)}
=2.948\times10^{22}\times5.21\times10^{-3}
\approx1.54\times10^{20}$~cm$^{-3}$.\\
\texttt{\textless /think\textgreater}\\[0.35em]
Final answer: $\boxed{1.54\times10^{20}\ \mathrm{defects/cm^3}}$
(reference answer: $1.5\times10^{20}$)
\end{OmniModelBox}

\Needspace{10\baselineskip}
\CaseTitle{Specific strength and embodied energy of Al vs.\ Ti alloys
(MatSciBench)}

\begin{OmniUserBox}
Aluminum alloy 7075-T6 (yield strength $Y_{\mathrm{Al}}=511$~MPa, density
$\rho_{\mathrm{Al}}=2.7$~g/cm$^3$) and Ti-6Al-4V (yield strength
$Y_{\mathrm{Ti}}=939$~MPa, density $\rho_{\mathrm{Ti}}=4.5$~g/cm$^3$) are
being considered for a 200~mm long tensile member carrying a load of 50~kN.
Assume a factor of safety of 1.5. Which material results in a lighter tensile
member, and which material results in a lower embodied-energy impact? Use the
specific embodied energies 193.5~MJ/kg (Al) and 915~MJ/kg (Ti). Give your
answer as a tuple $(m_{\mathrm{Al}},m_{\mathrm{Ti}},E_{\mathrm{Al}},
E_{\mathrm{Ti}})$ with units (g, g, MJ, MJ).
\end{OmniUserBox}

\begin{OmniModelBox}
\texttt{\textless think\textgreater}\\
(1)~Required cross-sectional areas:
$A=F/(Y/1.5)$, giving $m=\rho A L$ per material:
$m_{\mathrm{Al}}=79.26$~g, $m_{\mathrm{Ti}}=71.89$~g.\\
(2)~Sanity check on the mass ratio:
$m_{\mathrm{Al}}/m_{\mathrm{Ti}}
=(\rho_{\mathrm{Al}}/Y_{\mathrm{Al}})/(\rho_{\mathrm{Ti}}/Y_{\mathrm{Ti}})
=(2.7/511)/(4.5/939)=1.102$. ``Wait. Let's recheck the mass ratio. [...] My
mental math was wrong because I didn't divide by the yield strength in the
ratio. Aluminum is only about 10\% heavier.'' \\
(3)~Embodied energies:
$E_{\mathrm{Al}}=193.5\times0.07926=15.34$~MJ,
$E_{\mathrm{Ti}}=915\times0.07189=65.77$~MJ, so aluminum is
considerably lighter in embodied energy.\\
\texttt{\textless /think\textgreater}\\[0.35em]
Final answer: $\boxed{(79.26,\ 71.89,\ 15.34,\ 65.77)}$
(reference answer: $(79.3,\ 71.9,\ 15.4,\ 65.7)$)
\end{OmniModelBox}

\end{CJK*}
\end{document}